\PassOptionsToPackage{breaklinks,pagebackref, colorlinks}{hyperref}
\documentclass[]{fairmeta}

\usepackage{amsmath,amsfonts,bm}

\def\eqref#1{equation~\ref{#1}}
\def\Eqref#1{Equation~\ref{#1}}
\def\1{\bm{1}}

\def\vzero{{\bm{0}}}
\def\vone{{\bm{1}}}
\def\vmu{{\bm{\mu}}}

\def\va{{\bm{a}}}

\def\vm{{\bm{m}}}

\def\vs{{\bm{s}}}

\def\vu{{\bm{u}}}
\def\vv{{\bm{v}}}

\def\vx{{\bm{x}}}

\DeclareMathAlphabet{\mathsfit}{\encodingdefault}{\sfdefault}{m}{sl}
\SetMathAlphabet{\mathsfit}{bold}{\encodingdefault}{\sfdefault}{bx}{n}

\DeclareMathOperator*{\argmax}{arg\,max}

\usepackage[utf8]{inputenc}
\usepackage[T1]{fontenc}
\usepackage{microtype}

\usepackage[dvipsnames,table]{xcolor}
\definecolor{okabeBlue}{HTML}{0072B2}
\definecolor{okabeGreen}{HTML}{009E73}
\definecolor{okabeRed}{HTML}{D55E00}
\definecolor{okabeYellow}{HTML}{F0E442}
\definecolor{okabePurple}{HTML}{CC79A7}
\definecolor{hkuBlue}{HTML}{04B2D9}
\definecolor{hkuGreen}{HTML}{03A678}
\usepackage{amsmath}
\usepackage{amsfonts}
\usepackage{amssymb}
\usepackage{amsthm}
\usepackage{mathtools}
\usepackage{bm}
\usepackage{nicefrac}
\usepackage{dsfont}
\usepackage{bbm}
\usepackage{graphicx}
\usepackage{svg}
\usepackage{booktabs}
\usepackage{multirow}
\usepackage{makecell}

\usepackage{tabularx}
\usepackage{threeparttable}
\usepackage{caption}
\usepackage{subcaption}
\usepackage{wrapfig}
\usepackage{needspace}
\usepackage{rotating}
\usepackage{float}
\usepackage{dblfloatfix}
\usepackage[shortlabels]{enumitem}
\usepackage{url}
\usepackage{siunitx}

\usepackage{tcolorbox}
\usepackage{pifont}

\usepackage{pgfplots}
\pgfplotsset{compat=1.18}
\usepgfplotslibrary{groupplots}

\usepackage[ruled,vlined,linesnumbered]{algorithm2e}
\DontPrintSemicolon
\SetAlgoNoEnd
\SetAlFnt{\footnotesize}
\SetAlgoInsideSkip{}
\SetKwProg{Def}{function}{}{}
\SetKwFor{For}{for}{do}{}
\SetKwIF{If}{ElseIf}{Else}{if}{then}{else if}{else}{}
\SetKw{In}{in}

\SetCommentSty{algcomment}
\SetKwComment{Comment}{$\triangleright$\,}{}
\usepackage{listings}
\usepackage{fancyvrb}
\newcommand{\eg}{\emph{e.g.}}
\newcommand{\ie}{\emph{i.e.}}

\newcommand{\method}{\textsc{VPFlow}}
\newcommand{\lambdastate}{\lambda_{\mathrm{state}}}

\newcommand{\epsv}{\bm{\varepsilon}}
\newcommand{\vsigma}{\bm{\sigma}}
\newcommand{\epses}{\mbox{$\varepsilon$-ES}}
\newcommand{\epscem}{\mbox{$\varepsilon$-CEM}}
\newcommand{\acem}{\mbox{$\mathcal{A}$-CEM}}
\newcommand{\epssgd}{\mbox{$\varepsilon$-SGD}}
\newcommand{\epsadam}{\mbox{$\varepsilon$-Adam}}

\newcommand{\Hexec}{H_{\mathrm{exec}}}
\newcommand{\Nelite}{N_{\mathrm{e}}}

\newenvironment{maintextlayout}{%
  \titlespacing*{\subsection}{0pt}{1.8ex plus 0.4ex minus 0.2ex}{0.8ex plus 0.2ex}%
  \setlength{\floatsep}{6pt plus 2pt minus 2pt}%
  \setlength{\intextsep}{8pt plus 2pt minus 2pt}%
}{}

\hypersetup{
  linkcolor=Red,
  citecolor=Blue,
  urlcolor=Magenta,
}     % hyperlinks

\title{\textcolor{okabeBlue}{V}\textcolor{okabeGreen}{P}Flow: Value-Predictive Planning with Imperfect World Models for Offline Goal Reaching}
\author{Yunzhe Hu, Dong Xu}
\affiliation{School of Computing and Data Science \\ The University of Hong Kong}

\abstract{Offline goal-conditioned reinforcement learning enables skill acquisition by reaching any goal from any state without dense rewards. Model-based methods allow for planning by scaling test-time compute to solve complex tasks. Combining both holds appealing promise for pre-training on suboptimal demonstrations and zero-shot adaptation via goal specification and planning. Yet existing methods assume effective planning requires accurate next-state prediction in pixel or latent space. In this work, we show that value prediction is more useful for planning, whereas accurate state prediction is often unnecessary. Our key insight is that offline goal reaching naturally admits a closed-form Monte Carlo return, enabling value regression to replace temporal-difference bootstrapping. We use this return as the sole supervision for latent world models: imagined latents are trained to predict goal-reaching value rather than match future states, implicitly preventing representation collapse. These planning-oriented representations let imperfect world models imagine on the level sets of value field beyond dataset transitions. We then present \method{}, a value-predictive planner that pairs the world models with a one-step flow policy and plans in its noise space via evolutionary search. It enables policy improvement with imagined generalized advantage and reduces out-of-distribution queries common in action-space planning. On 30 state and pixel OGBench datasets, \method{} surpasses model-free baselines by at least 30.3 and 10.1 percentage points in success rate, respectively, and outperforms latent planners LeWorldModel, PLDM, and TD-MPC2. It achieves this despite a latent prediction error orders of magnitude larger than with state prediction, while minimizing this error lowers success rates. This demonstrates that value-predictive representations, not exact state prediction, are what offline planning requires.

}
\correspondence{\email{yzhu@cs.hku.hk}}
\begin{document}

\maketitle

% \begin{maintextlayout}
%%%%%%%%%%%%%%%%%%%%%%%%%%%%%%%%%%%%%%%%%%%%%%%%%%%%%%%%%%%%
\section{Introduction}
\label{sec:intro}
% First paragraph: introduce the offline goal reach (or offline goal-conditioned RL) problem, why it
% is important; then introduce the benefits of model-based methods in test-time planning; elucidate
% why combining both is challenging, less explored, yet promising: pre-training with offline data
% and adapting to new goals in zero-shot via test-time planning
Offline reinforcement learning (RL) seeks to learn reusable behaviors purely from previously
collected trajectories with no additional environment interaction \citep{levine2020offline}.
In its goal-conditioned setting, an agent is tasked with reaching a desired goal at test time without optimizing a hand-designed reward \citep{kaelbling1993learning,schaul2015universal, park2025ogbench}.
This makes offline goal reaching especially attractive for reward-free skill acquisition: relabeling future observations as goals turns one unlabeled dataset into supervision for many tasks \citep{andrychowicz2017hindsight}.
Pairing it with model-based planning is a natural next step: pre-train on offline data once, then adapt to new goals in zero shot by searching over imagined futures at test time.
This combination is promising, yet comparatively unexplored.

\begin{figure}[t]
  \centering
  \includegraphics[width=\textwidth]{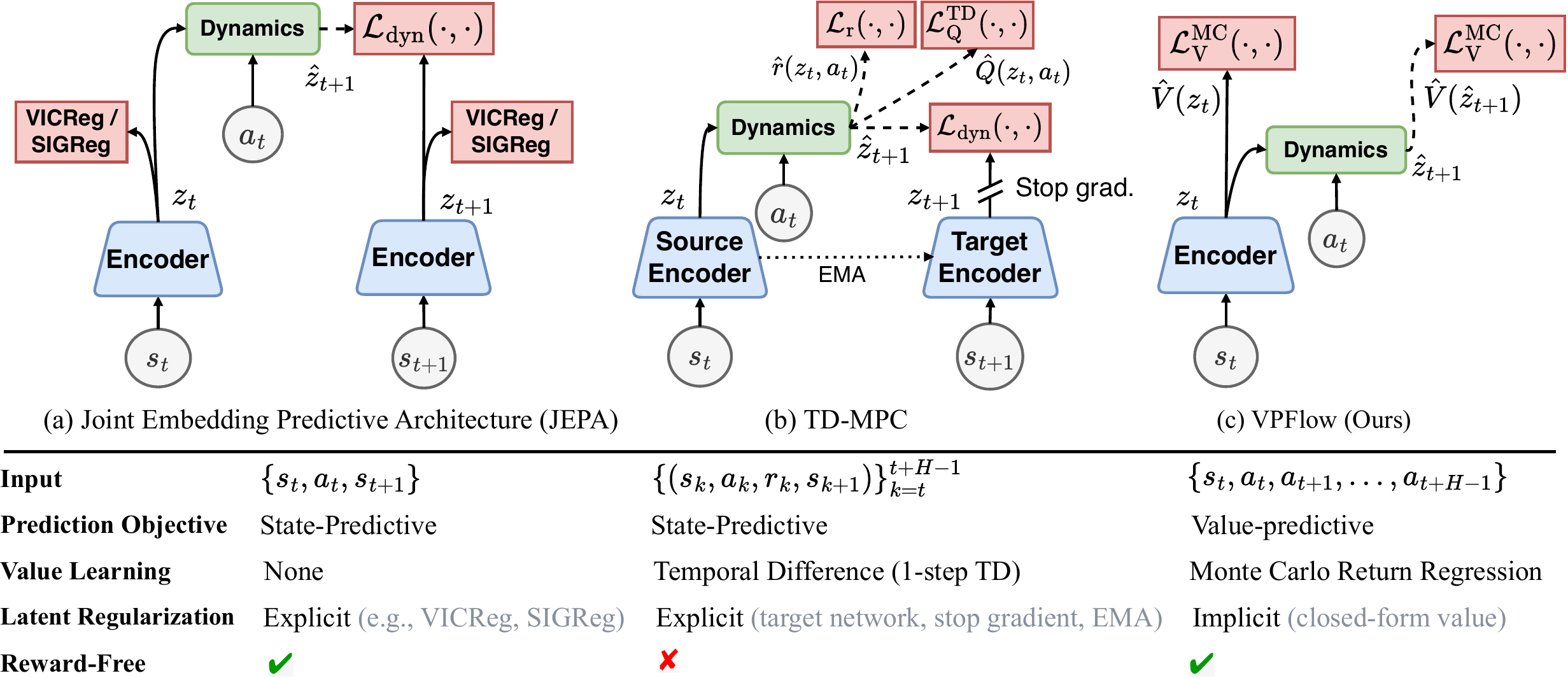}
  \caption{\textbf{State-predictive and value-predictive action-conditioned latent world models.}
    \emph{Left}: Joint-embedding predictive architecture (JEPA) aligns predicted and future
    latents with a joint encoder and explicit regularization, \eg, variance-invariance-covariance
    regularization (VICReg) or sketched isotropic Gaussian regularization (SIGReg).
    \emph{Middle}: TD-MPC learns values from one-step
    temporal-difference (TD) targets, with source and target encoders, stop-gradient, and
    exponential moving average (EMA) heuristics.
    \emph{Right}: \method{} trains the latent dynamics and encoders jointly via value prediction:
    predicted latent states seek accurate goal-reaching value estimates, not future states.
    Closed-form hindsight Monte Carlo (MC) returns provide the temporal signal and act as an implicit regularizer.}
    % eliminating the need for EMA and target encoders.
  \label{fig:architecture}
\end{figure}
% Second paragraph: how existing model-based methods (e.g., LeWorldmodel, PLDM, TD-MPC) approach
% this problem setting, from pure state/image space to latent space; state their key assumptions and
% why it fall short, what is missing
Strong offline goal-conditioned methods, spanning implicit Q-learning \citep{kostrikov2022offline}, contrastive learning \citep{eysenbach2022contrastive}, and quasi-metric representations \citep{wang2023qrl,myers2025quasimetric}, learn a reactive policy or value function from relabeled
trajectories and act greedily at deployment.
While robust and efficient, they are static at test time: once training ends, additional inference compute cannot improve decisions on challenging tasks.
Model-based planning offers a complementary advantage.
From Dyna to TD-MPC \citep{sutton1991dyna,hafner2019learning,hafner2020dream,pmlr-v162-hansen22a}, learned world models convert extra test-time compute into better decisions by searching over imagined futures.
Yet these methods share a foundational assumption that has migrated from pixels to latents largely unquestioned: the world model must accurately predict the next state, or the next latent embedding,
for planning to succeed \citep{hafner2019learning,schwarzer2021dataefficient,sobal2025pldm,maes2026leworldmodel}.
% Offline goal reaching is the setting where this assumption is most fragile. Trajectories are
% sparse and heterogeneous, so a state-predictive model accumulates error over the very horizons on
% which planning is supposed to help, and no online data arrives to correct it.

% Third paragraph: state our key insight and argument; discuss its motivation and connections with
% the most relevant prior work; delineate the big picture; and finally identify the objective and
% core question: the useful representation for planning \todo{introduce and discuss the
% value-equivalence view of model-based control, as well as value-prediction networks, universal
% value function, predictive/successor representations, and finally transition to value-predictive
% representations.}

We argue that this state-predictive assumption is both overly restrictive and misdirected.
What a planner needs is not a model that reproduces the next observation, but one that preserves the
relative ordering of candidate futures according to their goal-reaching value.
This view builds on a rich lineage: value prediction networks learn dynamics only as much as they
reproduce Bellman backups \citep{oh2017vpn}; universal value functions and successor representations
factor value into future occupancy and a goal-dependent readout
\citep{dayan1993improving,schaul2015universal,barreto2017successor}; and the value-equivalence
principle formalizes when a model that is inaccurate as a simulator remains exact for control
\citep{grimm2020value}.
We take this principle a step further: an imagined latent only needs to land on the right level set
of the value field, so the representation worth learning for planning is \emph{value-predictive} rather than \emph{state-predictive}.
Offline goal reaching makes this directly actionable.
Because the relabeled reward is sparse, the Monte Carlo (MC) return from any state to any future
goal on its trajectory is a closed-form function of their temporal distance.
Value learning thus simplifies from temporal-difference bootstrapping to direct regression, and
that regression target can supervise the latent dynamics end to end, acting as an implicit
regularizer against representation collapse.
A model trained this way may be an imperfect simulator yet remain an effective planner.

% Fourth paragraph: introduce how to leverage this view and our methodologies that are informed by
% it
Motivated by this view, we introduce \textbf{Value-Predictive Flow (\method{})}, which plans for offline goal-reaching tasks with a deliberately imperfect world model. 
\method{} jointly learns an observation encoder, a goal-agnostic latent dynamics model, and goal-conditioned value heads alongside a distilled flow-matching policy with action chunks.
Imagined latents are trained purely to be \emph{value-predictive}: rolled forward with dataset
actions, the value predicted at each imagined latent is regressed towards the closed-form return of
the corresponding future state.
No observation is reconstructed, no target latent is matched, and no bootstrapped target network is
required.
Figure~\ref{fig:architecture} contrasts alternative latent-predictive architectures with ours.

% Fifth paragraph: similar to the fourth paragraph, but focus on the test-time planning in the noise
% space with the generalized advantage function; provide insights on noise steering that can largely
% avoid the OOD action queries; name the planning as \epscem{}, \epssgd{} and \epsadam{}
\looseness-1
To translate a value-predictive model into reliable control, the planner must propose feasible,
in-support actions.
Standard zeroth-order planners such as the cross-entropy method (CEM) and model predictive path
integral (MPPI) \citep{rubinstein1999cem,williams2017mppi} perturb raw action sequences, which in
offline settings quickly leave the data support and expose the value function to out-of-distribution
(OOD) queries.
\method{} instead searches in the noise space of its one-step flow policy \citep{park2025flow},
where each candidate noise vector decodes into action chunks at the current latent state.
Because of behavior-cloning regularization, the noise search warm-starts at multi-modal, in-support
proposals and stays anchored to the data manifold.
We guide this noise steering using the generalized advantage estimator (GAE) \citep{schulman2016gae}
over imagined rollouts, instantiated as an evolution strategy via annealed mutation and
elite selection.

% Sixth paragraph: state the main results, evaluation and findings; focus on the most valuable and
% interesting ones, not just a detailed case
Across 30 locomotion and manipulation datasets from OGBench \citep{park2025ogbench}, \method{}
raises the mean success rate from $32.9\%$ to $63.2\%$ over 17 state datasets and from $27.5\%$ to
$37.6\%$ over 13 pixel datasets relative to the strongest non-hierarchical baselines, while
substantially outperforming the latent state-predictive planners LeWorldModel \citep{maes2026leworldmodel}, PLDM \citep{sobal2025pldm}, and TD-MPC2 \citep{hansen2024td}.
Crucially, these gains hold while the model's next-latent prediction error sits orders of magnitude
above that of its state-predictive counterpart; conversely, enforcing state prediction closes that
gap at the cost of lowering the success rate.
This demonstrates that offline planning does not require a faithful state simulator, but
rather a value-predictive one. Our contributions are as follows:
% not an optimal design choice we commonly believed.
\begin{enumerate}[leftmargin=*,itemsep=1pt,parsep=0pt,topsep=2pt]
  \item We argue and demonstrate that value-predictive latent world models, trained solely via
    closed-form Monte Carlo return regression, avoid representation collapse and suffice for offline
    planning even as state-prediction error grows, challenging the pervasive assumption of
    learning state-predictive world models for planning.

  \item We propose a test-time planning algorithm tailored to generative flow policies in their
    noise space, steering initial noise towards higher imagined generalized advantage while avoiding
    the out-of-distribution action queries of existing action-space planning methods.

  \item We show across 30 state and pixel datasets on OGBench that \method{} surpasses model-free
    baselines and latent state-predictive planners, and scales favorably with test-time compute on
    long-horizon tasks.
\end{enumerate}

% \begin{enumerate}[leftmargin=*] \item We show that the state-predictive requirement on world
% models is unnecessary for offline goal reaching, and give a controlled objective sweep in which
% value-predictive latent dynamics, trained purely by closed-form Monte Carlo regression, support
% planning while observation-prediction error increases.

%   \item We formulate test-time search as steering a distilled flow policy in its own noise space,
%   and give two optimizers over that space, \epscem{} and gradient ascent, that share one objective
%   and differ only in how they use derivative information.

%   \item We evaluate on the full OGBench state and pixel suites, report where each steering variant
%   wins and why, and provide diagnostics that expose the values, advantages, and noise-space
%   populations the controller actually optimizes. \end{enumerate}

\begingroup
% Keep the section on page 3 with local spacing only.
% \setlength{\parskip}{2pt plus 1pt minus 1pt}
\titlespacing*{\section}{0pt}{2.4ex plus 0.5ex minus 0.2ex}{1.4ex plus 0.2ex}
\section{Related Work}
\label{sec:related}

We position \method{} within three lines of work with broader literature in
Appendix~\ref{app:related}.

% \looseness-1
\textbf{Offline goal-conditioned RL.}
Offline RL learns a policy from pre-collected datasets without interacting with the environment,
and needs techniques to keep actions on the data manifold during value learning
\citep{levine2020offline,kostrikov2022offline,
  kumar2020conservative,fujimoto2021minimalist,park2025flow}. In the goal-conditioned setting, an
agent aims to reach any goal from any state \citep{kaelbling1993learning}, and hindsight
relabeling \citep{andrychowicz2017hindsight} can convert any trajectories, even failed ones, into
successful ones with the goal actually reached. With high-coverage datasets, this may enable
learning a universal value function that empowers the policy to generalize to unseen goals in
zero shot \citep{schaul2015universal,park2024hilp}.
% \todo{add references for wang2023qrl,myers2025quasimetric, eysenbach2022contrastive; one sentence only.}

\looseness-1
\textbf{Representations for planning with world models.}
World models seek a compact representation and transition dynamics from observations, where value
can be learned in imagination \citep{sutton1991dyna,ha2018world,hafner2020dream,pmlr-v162-hansen22a}. Yet what representations are more effective for planning is
still an open question. Reconstruction-based methods harness the dense supervision from pixels and
have achieved remarkable success \citep{hafner2019learning,hafner2025dreamerv3}. Self-predictive
representation holds promise for a more powerful model
\citep{schwarzer2021dataefficient,lecun2022path,bagatella2026tdjepa}, but needs stop-gradient, EMA,
or regularization to prevent collapse \citep{sobal2025pldm,maes2026leworldmodel}. More critically,
it fails to exploit the rich temporal-distance supervision naturally available from goal reaching.
This makes its usefulness for planning not directly obvious. Our work aims to reveal what is
essential and necessary for planning-oriented representations.

% \looseness-1
\textbf{Test-time search.}
With a learned world model, previous studies primarily plan actions by sampling-based optimization
\citep{williams2017mppi,janner2019mbpo,hafner2019learning,pmlr-v162-hansen22a, park2026mac} or
search \citep{schrittwieser2020muzero}. Yet action-space planning may suffer heavily in the offline
setting, especially with low-coverage data, in that off-manifold actions can severely exploit the
planning objective.
Advances in generative policies have turned to steering the noise at test time
to keep actions in-distribution, but they need reward or other external signals and have yet to
show scaling effectiveness \citep{yang2024diffusiones,wagenmaker2025dsrl,patil2026goldenticket}. 
In contrast, \method{} can improve itself at test time by noise steering\footnote{We use ``steering'', ``planning'', and ``searching'' interchangeably in this paper.} with imagined
rollouts, even with imperfect world models.
% \par
\endgroup

% \end{maintextlayout}

% Keep the class spacing for Preliminaries and Method.
\section{Problem Statement and Preliminaries}
\label{sec:prelim}

\textbf{Offline goal reaching.}
We consider the offline goal-conditioned reinforcement learning problem as a reward-free controlled
Markov decision process (MDP) $\mathcal{M} = (\mathcal{S}, \mathcal{A}, \gamma, P)$, where
$\mathcal{S}$ denotes the state space, $\mathcal{A}$ denotes the action space, $\gamma \in [0,1)$
denotes the discount factor, and $P: \mathcal{S} \times \mathcal{A} \to \Delta(\mathcal{S})$
denotes the transition dynamics mapping state and action to another state distribution
$\Delta(\cdot)$. Given a dataset $\mathcal{D}=\{\tau^{(i)}\}_{i=1}^{N_{\mathcal{D}}}$ of
$N_{\mathcal{D}}$
trajectories $\tau^{(i)}=(s_0^{(i)},a_0^{(i)},s_1^{(i)},\dots,s_{T_i}^{(i)})$ of length $T_i$,
collected by a behavior policy\footnote{No assumption on its optimality.}, the objective is
to learn a goal-conditioned policy $\pi(a \mid s, g)$ that maximizes the expected discounted
utility $\mathbb{E}_{\tau}[\sum_{t=0}^{\infty} \gamma^t \delta_g(s_t)]$ for any goal $g\in\mathcal{S}$. Here, $\delta_g(s_t)$ has been referred to as the
``Dirac delta function'' \citep{park2025ogbench}. In practice, we consider it as the indicator
function $\mathds{1}_{\{s_t = g\}}$ that represents the relabeled sparse reward:
% \vspace{-0.6em}
\begin{equation}
  \label{eq:sparse_reward}
  r(s_t, g) \coloneqq
  \begin{cases}
    1, & \text{if } s_t = g,          \\
    0, & \text{otherwise}.
  \end{cases}
\end{equation}
\textbf{Model predictive control.}
Given a transition dynamics model $f: \mathcal{S} \times \mathcal{A} \to \mathcal{S}$, model
predictive control (MPC) addresses an online constrained optimal control problem with a lookahead
horizon $H$ and an action sequence $a_{t:t+H-1}=(a_t,\dots,a_{t+H-1})$ at the current state $s_t$,
then executes the first action $a_t$, and re-plans at the next state. More formally, MPC solves the
following optimization problem at time step $t$:
% \vspace{-0.6em}
\begin{equation}
  \label{eq:mpc}
  \min_{a_{t:t+H-1}} \sum_{h=0}^{H-1} \ell_t(s_{t+h}, a_{t+h}) + \ell_f(s_{t+H}), \quad \text{s.t.}
  \quad s_{t+1} = f(s_t, a_t),
\end{equation}
where $\ell_t(s, a)$ and $\ell_f(s)$ are the running and terminal cost functions of interest. After
finding the minimizer $a_{t:t+H-1}^{\star}$ of \Eqref{eq:mpc}, only the first action $a_t^{\star}$
is applied due to the compounding error in the dynamics.
\looseness-1
In the model-based RL context, the cost has been instantiated using negative reward $r_t(s_t, a_t)$ and the value function $Q(s_t, a_t)$ or $V(s_t)$. Both are learned as part of
the world model $f(s_t, a_t)$, which can be used to plan without interacting with the real
environment \citep{hafner2019learning,hafner2020dream,pmlr-v162-hansen22a}. Planning is widely done by sampling-based trajectory optimizers \citep{rubinstein1999cem,williams2017mppi}.
Another emerging way is to learn planning-enabled representations without the reward and value
function, and use a simple $\ell_2$ distance to the goal as the cost
\citep{assran2025v,sobal2025pldm,maes2026leworldmodel}. 
Both lines of work assume the optimal objective is to predict the next state when learning world models. In this work, we provide evidence for another value-predictive representation view and show that it is more useful for planning.
% provide an alternative perspective on the effective representation for planning.

\section{VPFlow: Planning with Imperfect World Models}
\label{sec:method}
% \vspace{-0.5em}
% \method{} learns an observation encoder $E_\phi$, latent dynamics $F_\phi$, twin goal-conditioned
% value heads $V_{\phi,1},V_{\phi,2}$, and a flow policy $\pi_\theta$ over action chunks.
% Sections~\ref{sec:mc}--\ref{sec:vpr} settle what the latent must preserve, which is goal-reaching
% value rather than the next state, and Sections~\ref{sec:flow-policy}--\ref{sec:steering} settle
% where the search lives, which is the noise space of a distilled flow policy.
\subsection{Value Equivalence Principle}
\label{sec:value-equivalence}
\begin{wrapfigure}[10]{r}{0.5\textwidth}
  \centering
  \vspace{-24pt}
  \includegraphics[width=0.48\textwidth]{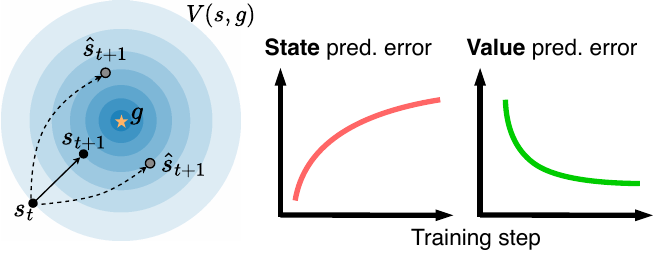}
  \caption{\textbf{The value equivalence principle view of world modeling.} The value-predictive
    representations of the same timestep constitute a level set of the value field.}
  \label{fig:value-equivalence}
\end{wrapfigure}

We motivate \method{} through the value equivalence principle \citep{grimm2020value}, which states
that two \emph{dynamics and reward models} are value-equivalent with respect to a set of value functions if they induce the same Bellman operator. Consequently, one can learn another set of dynamics and reward models beyond the training examples but still keep the same effect for policy learning.

We extend this view to world modeling for offline goal reaching: two \emph{predictive representations} $\hat{s}, \hat{s}' \in \mathcal{S}$ are value-equivalent with respect to a goal if they are evaluated identically by the goal-conditioned value function, \ie, $V_\phi(\hat{s}, g) = V_\phi(\hat{s}', g)$ (Figure~\ref{fig:value-equivalence}). 
In this setting, the Bellman operator can be written in closed form with sparse reward given a goal state, reducing recursive temporal-difference learning \citep{oh2017vpn,schrittwieser2020muzero} to direct regression on fixed targets. 
Intuitively, rather than constraining the dynamics models to match the future states, this principle only requires the predicted representations to lie on the corresponding level sets of value functions, and this relaxation can potentially expand the reachable space of dynamics models.
% We provide a formal characterization in Appendix~\ref{app:theory}.

\subsection{Hindsight Monte Carlo Return Supervision}
\label{sec:mc}
\begin{figure}[tbp]
  \centering
  \includegraphics[width=\textwidth]{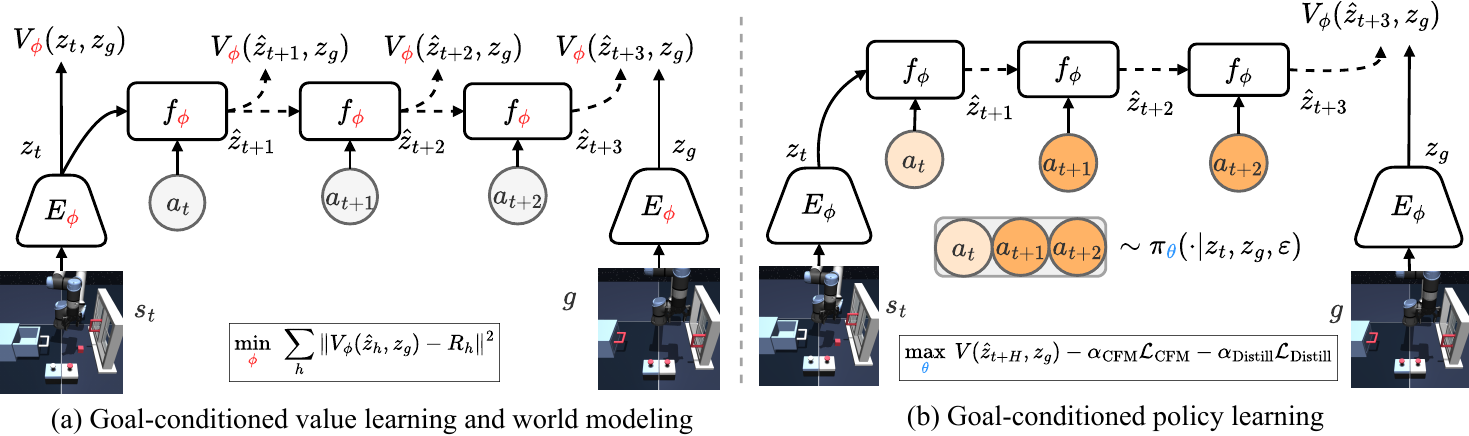}
  \caption{\textbf{\method{} objectives.}
    \method{} alternately trains (a) the goal-conditioned value function $V_\phi$,
    latent dynamics $f_\phi$, and encoder $E_\phi$ jointly, and (b) a distilled one-step flow policy
      $\pi_\theta$.
    The value function regresses the closed-form Monte Carlo returns $R_h$ on imagined rollouts. The
    policy is improved by maximizing the imagined value from the learned dynamics model, regularized
    by conditional flow matching to stay close to the
    data manifold.}
  \label{fig:vpflow-objective}
\end{figure}

Q-learning typically uses bootstrapping that predicts Bellman backup online with a target network
and EMA to learn the action-value function $Q(s,a)$. Though this heuristic stabilizes training
\citep{tang2023understanding}, it makes TD learning less principled for scaling on larger datasets
\citep{park2026horizon}.

With hindsight relabeling, every trajectory in the dataset can be relabeled with a goal as if the
agent in that trajectory has reached that goal. This naturally admits a cleaner and simpler
objective for value learning. Concretely, the target value for a state $s_t$ is the discounted
Monte Carlo return from $s_t$ to the relabeled goal $g=s_{t+d}$ on the same trajectory:
% \vspace{-0.8em}
\begin{equation}
  R(s_t, g) = R(d) = \sum_{k=0}^{d}\gamma^k r(s_{t+k}, g) = \begin{cases}
    \gamma^d,                      & r(s, g) =\mathds{1}_{\{s=g\}}     \\
    \frac{1-\gamma^{d}}{\gamma-1}, & r(s, g) =\mathds{1}_{\{s=g\}} - 1
  \end{cases},
  \label{eq:mc_target}
\end{equation}
where $d \geq 0$ is the temporal distance, \ie, the number of steps from $s_t$ to the goal $g$ on
the same trajectory. In practice, we use the sparse reward $r(s, g) =\mathds{1}_{\{s=g\}} - 1$,
following prior work \citep{intelligence2025pi, park2025ogbench}.
Intuitively, this quantity measures how far the agent is from the goal as a progress signal.
% In principle, it represents a sample of value from the behavior policy, maximizing it does not
% guarantee policy improvement. But with a sufficiently random trajectories that covers the optimal
% policy, it can be a good proxy as the real value despite potential high variance. \todo{is this
% claim true?}

The training of \method{} alternates between the value prediction and policy learning
objectives, as shown in Figure~\ref{fig:vpflow-objective} and elaborated below.
\subsection{Value-Predictive Latent Dynamics}
\label{sec:vpr}

Given this closed-form Monte Carlo return in \Eqref{eq:mc_target}, \method{} learns the observation
encoder $E_\phi: \mathcal{S} \to \mathcal{Z}$ for a latent space $\mathcal{Z}$, latent dynamics
$f_\phi$, and a goal-conditioned value function
$V_\phi: \mathcal{Z} \times \mathcal{Z} \to \mathbb{R}$. All three components are optimized
jointly with the value-prediction (VP) objective. In particular, we use the action chunk
representation of horizon $H$ as $\va_t = a_{t:t+H-1}$, which allows for a certain level of temporal
abstraction \citep{li2025reinforcement}.
After encoding the state and goal with $E_\phi$:
\begin{equation}
  z_t = E_\phi(s_t), \quad z_g = E_\phi(g),
\end{equation}
the latent dynamics predicts each future latent with an action prefix:
\begin{equation}
  \label{eq:latent_dynamics}
  \hat z_{t+h}=f_\phi(z_t,a_{t:t+h-1}), \quad h=1,\ldots,H, \quad \hat z_t\coloneqq z_t.
\end{equation}
Unrolling $f_\phi$ with the action chunk $\va_t$ yields intermediate latents $\hat z_{t+1},\dots,\hat z_{t+H}$, where $\hat z_{t+H}=f_\phi(z_t,\va_t)$ denotes the final predicted latent. The VP loss is then defined as:
\begin{equation}
  \mathcal{L}_{\mathrm{VP}}(\phi)
  \coloneqq
  \mathbb{E}_{(s_t, \va_t) \sim \mathcal{D}, g \sim \mathcal{S}} \left[ \frac{1}{H+1}\sum_{h=0}^{H}
  M_{t+h} \odot \|V_{\phi}(\hat z_{t+h},z_{g})-R_{t+h}\|_2^2 \right], \quad \hat z_t \coloneqq
  z_t,
  \label{eq:value_prediction_loss}
\end{equation}
where $R_{t+h}=R(s_{t+h},g)$ is the target of \Eqref{eq:mc_target} for the dataset state $s_{t+h}$
and does not depend on $\phi$, $\odot$ is the Hadamard product, and the mask $M_{t+h}\in\{0,1\}$
filters out goals behind time $t+h$ and across trajectories such that the value function only
exploits the clean temporal distance signal on the same trajectory.

% \todo{the state predictive loss variant should resort to this equation and with state prediction
% loss; explain in appendix}

\Eqref{eq:value_prediction_loss} trains the encoder, latent dynamics, and value function jointly.
The value supervision nudges the predictive representation of each rollout horizon towards the level
set of the value function defined by the state-goal temporal distance, instead of aiming for perfect
state prediction. It structures the representation space with explicit meaning as progress towards
reaching the goal, unlike the regularizers that are only statistically meaningful
\citep{bardes2022vicreg,balestriero2025lejepa}. It also expands the hypothesis space for the policy
without being too conservative inside the data manifold, as we discussed in
Section~\ref{sec:value-equivalence}.

\subsection{Flow Policy as a Behavior Prior}
\label{sec:flow-policy}

For a more expressive policy to handle the multimodality of offline datasets, we opt for a
flow-based policy similar to Flow Q-learning \citep{park2025flow}. Beyond that, the
generative policy as a behavior prior is also a key design choice to enable planning with the data
manifold constraint at test time.

The policy improvement objective is to maximize the Q-function of an action chunk
\citep{li2025reinforcement} given encoded state and goal, $z_t = E_\phi(s_t), z_g = E_\phi(g)$,
using the learned world models:
\begin{equation}
  \begin{aligned}
    Q(z_t, \va_t, z_g) & \coloneqq \mathbb{E}_{\hat z_{t+H} \sim f_\phi(z_t,
    \va_t)}\left[\sum_{h=0}^{H-1}\gamma^h r(z_{t+h}, z_g) + \gamma^H V_\phi(\hat z_{t+H},
    z_g)\right]
                       \\
                       & \approx \mathbb{E}_{\hat z_{t+H} \sim f_\phi(z_t, \va_t)}\left[\gamma^H
                         V_\phi(\hat z_{t+H},
      z_g)\right] \qquad \text{(no $\theta$ in reward)},
  \end{aligned}
  \label{eq:q_function}
\end{equation}
where the action chunk $\va_t = a_{t:t+H-1}$ is decoded from Gaussian noise
$\epsv\sim\mathcal{N}(\vzero,I)$ by the one-step flow policy
$\pi_\theta(z_t, z_g, \epsv)$ with parameters $\theta$. Together with a velocity field
$v_\theta$, the policy learning loss is defined as:
\begin{equation}
  \mathcal{L}_{\pi}(\theta)
  \coloneqq
  - Q(z_t, \va_t, z_g)
  +\alpha_{\mathrm{CFM}}\mathcal{L}_{\mathrm{CFM}}
  +\alpha_{\mathrm{Distill}}\mathcal{L}_{\mathrm{Distill}} .
  \label{eq:actor_loss}
\end{equation}
Here, $\mathcal{L}_{\mathrm{CFM}}$ is the conditional flow matching (CFM) loss that fits $v_\theta$
to the dataset actions, $\mathcal{L}_{\mathrm{Distill}}$ distills the multi-step decoding process
using $v_\theta$ into $\pi_\theta$, and $\alpha_{\mathrm{CFM}},\alpha_{\mathrm{Distill}}$ are their
weights. The one-step policy makes noise steering more efficient.
Detailed objective functions are explained in Appendix~\ref{app:objectives}.

\subsection{Steering Noise as Planning}
\label{sec:steering}

\textbf{Planning objective.}
\method{} instantiates the cost objective in \Eqref{eq:mpc} with the reward function $r_t$ defined
in \Eqref{eq:sparse_reward} and optimizes the action chunk $\va_t = a_{t:t+H-1}$ at state $z_t$.
Yet, credit assignment is challenging for sparse reward. As such, we densify it with the
potential-based reward shaping \citep{ng1999policy} using imagined rollouts, which yields the
reshaped reward as:
% \vspace{-0.8em}
\begin{equation}
  \delta_t
  =
  r_t+\gamma V_\phi(\hat{z}_{t+1},z_g)-V_\phi(\hat{z}_t,z_g).
  \label{eq:reshaped_reward}
\end{equation}
We design the planning objective to maximize the $(\gamma\lambda)$-discounted sum of the reshaped
reward over the planning horizon $H$, where $\lambda\in[0,1]$ is the trace-decay coefficient:
% \vspace{-0.3em}
\begin{equation}
  A(z_t, \va_t, z_g) = A_t \coloneqq  \sum_{h=0}^{H-1} (\gamma \lambda)^h \delta_{t+h}, \quad
    \text{s.t.} \quad \hat
  z_{t+h}=f_\phi(\hat{z}_t,a_{t:t+h-1}), \quad \hat{z}_t \coloneqq z_t.
    \label{eq:planning_objective}
\end{equation}
\Needspace{30\baselineskip}
\begin{wrapfigure}[27]{r}{0.58\textwidth}
  \SetAlCapFnt{\footnotesize}
  \SetAlCapNameFnt{\footnotesize}
  \SetAlgoSkip{}
  \setlength{\algomargin}{1em}
  \SetInd{0.35em}{0.65em}
\begin{algorithm}[H]
  \SetVlineSkip{0.3ex}
  \caption{\epses{}: evolutionary search in noise space with the generalized advantage estimator.}
  \label{alg:steering}
  \KwIn{state $s_t$, goal $g$; chunk length $H$; number of chunks $n_c$; elites $1\le\Nelite\le N$;
    population $N$; iterations $K\ge0$; mutation scale bounds
    $\sigma_{\min},\sigma_{\max}$; decay $\rho$; inverse temperature $\beta$; execution length
      $\Hexec$}
  \KwOut{first $\Hexec$ actions ($[{:}\Hexec]$ denotes slicing)}
  $z_t\leftarrow E_\phi(s_t)$, $z_g\leftarrow E_\phi(g)$\;
  \Def{$\operatorname{Dec}(\epsv=(\epsv_0,\ldots,\epsv_{n_c-1}))$}{
    $\hat z_t\leftarrow z_t$\;
    \For{$c=0,\ldots,n_c-1$}{
      $\va_c\leftarrow\operatorname{clip}(\pi_\theta(\hat z_{t+cH},z_g,\epsv_c),-1,1)$\;
      $\hat z_{t+(c+1)H}\leftarrow f_\phi(\hat z_{t+cH},\va_c)$
        \Comment*[r]{\Eqref{eq:latent_dynamics}, latent rollouts}
    }
    \KwRet concatenation of $(\va_0,\ldots,\va_{n_c-1})$\;
  }
  \If{$K=0$}{
    \KwRet $\operatorname{Dec}(\vzero)[{:}\Hexec]$ \Comment*[r]{No search}
  }
  $A_0\leftarrow A(z_t,\operatorname{Dec}(\vzero),z_g)$ \Comment*[r]{\Eqref{eq:planning_objective},
    GAE}
  $(\epsv^{(e)},A_e)\leftarrow(\vzero,A_0)$, $e=1,\ldots,\Nelite$ \Comment*[r]{Policy initialization
    at zero}
  \For{$k=0,\ldots,K-1$}{
    $\sigma_k\leftarrow\max(\sigma_{\min},\sigma_{\max}\rho^k)$ \Comment*[r]{Annealed mutation
      scale}
    \For{$e=1,\ldots,\Nelite$}{
      \For{$j=1,\ldots,N/\Nelite$}{
        $\tilde\epsv^{(j)}\sim\mathcal{N}(\epsv^{(e)},\sigma_k^2 I)$ \Comment*[r]{Mutate elites}
        $\tilde A_j\leftarrow A(z_t,\operatorname{Dec}(\tilde\epsv^{(j)}),z_g)$
          \Comment*[r]{Evaluate rollouts
          with GAE}
      }
      $j^\star\leftarrow\argmax_{j\in\{1,\dots,N/\Nelite\}}\tilde A_j$\;
      \If{$\tilde A_{j^\star} > A_e$}{
        $(\epsv^{(e)},A_e)\leftarrow(\tilde\epsv^{(j^\star)},\tilde A_{j^\star})$
          \Comment*[r]{Greedy
          improvement}
      }
    }
  }
  $w_e\leftarrow\exp(\beta A_e)/\sum_{j=1}^{\Nelite}\exp(\beta A_j)$, $e=1,\ldots,\Nelite$\;
  $e^\star\sim\operatorname{Categorical}((w_e)_{e=1}^{\Nelite})$ \Comment*[r]{Sample elite}
  \KwRet $\operatorname{Dec}(\epsv^{(e^\star)})[{:}\Hexec]$ \Comment*[r]{Receding-horizon execution}
\end{algorithm}
\end{wrapfigure}

It corresponds to the generalized advantage estimator (GAE) \citep{schulman2016gae} of horizon $H$.
The score $A$ measures predicted progress towards the goal relative to the current state.
We discuss this planning objective in Appendix~\ref{app:gae}.

\textbf{Noise space steering.} Planning has been widely conducted by directly optimizing
\emph{open-loop} actions on an unconstrained support.
For long-horizon rollouts, actions tend to drift away from the data distribution, making the
planning objective unreliable.
One of the key benefits of a generative policy is that after training, its
noise space can be optimized to search for better actions while still being close to the training
distribution \citep{yang2024diffusiones,wagenmaker2025dsrl}.
We adopt this strategy to steer the noise to optimize \Eqref{eq:planning_objective} that uses the
imagined states unrolled from the learned dynamics.

\looseness-1
\textbf{\epses{}: evolution strategy in noise space.}
CEM planners used in model-based methods maintain an elite set of actions, from which an online mean
and standard deviation are updated \citep{janner2019mbpo,pmlr-v162-hansen22a,maes2026leworldmodel}.
However, this design choice does not directly translate to the noise space, as their distribution
differs. Inspired by a previous gradient-free optimizer for diffusion noise
\citep{yang2024diffusiones}, we instead develop an evolution strategy that mutates samples of each
elite in noise space and greedily keeps the ones that score higher by the objective in
\Eqref{eq:planning_objective}. This allows the policy prior learned during training to improve
itself by adding more compute at test time without weight updates. We present the search process,
which runs for $K$ iterations, in Algorithm~\ref{alg:steering} with
more details in Appendix~\ref{app:eval}.

\begin{maintextlayout}
% \vspace{-0.5em}
\section{Experiments}
\label{sec:experiments}

We aim to address four questions in this section: \textbf{Q1}: How does \method{} compare with
model-free baselines and model-based methods that similarly plan with latent world models for offline goal-reaching tasks? \textbf{Q2}: Are state-predictive representations useful for planning, or do the value-predictive ones play a more essential role? \textbf{Q3}: Does test-time planning help, and does the search space matter more than the search algorithm? \textbf{Q4}: How does the learned latent actually behave during rollouts?

\textbf{Setup.}
We use the official OGBench datasets and tasks \citep{park2025ogbench}: 17 state-based datasets and
13 pixel-based datasets. The metric is the goal-reaching success rate ($\%$, $\uparrow$), averaged
over the five tasks within each dataset. Model-free baselines are the non-hierarchical methods
reported by OGBench: goal-conditioned behavioral cloning (\textbf{GCBC})
\citep{ghosh2021learning}, goal-conditioned implicit value learning (\textbf{GCIVL}) and
goal-conditioned implicit Q-learning (\textbf{GCIQL}) \citep{kostrikov2022offline}, quasi-metric RL
(\textbf{QRL}) \citep{wang2023qrl}, and contrastive RL (\textbf{CRL})
\citep{eysenbach2022contrastive}.
% \footnote{We omit HIQL \citep{park2023hiql}, which relies on a subgoal hierarchy, because
% \method{} is a flat controller.}
Model-based baselines are the latent planners \textbf{LeWorldModel (LeWM)}
\citep{maes2026leworldmodel}, \textbf{PLDM}
\citep{sobal2025pldm}, and \textbf{TD-MPC2} \citep{hansen2024td}. Implementation details are in Appendix~\ref{app:impl}, and hyperparameters are in Appendix~\ref{app:hparams}.

\begin{table}[t]
  \centering
  \small
  \caption{\textbf{Success rate ($\%$, $\uparrow$) on OGBench compared against model-free
      baselines.} Entries are mean success rates over the datasets in each environment, with
      the number of datasets in parentheses. The best and second best results in each row are in bold
    and underlined, respectively.}
  \label{tab:summary}
  \begin{tabular}{lllcccccc}
    \toprule
    \textbf{Input}         & \textbf{Task Type}            & \textbf{Environment} & GCBC             & GCIVL            & GCIQL            & QRL           & CRL              & \method{}        \\
    \midrule
    \multirow{6}{*}{State} & \multirow{2}{*}{Locomotion}
                           & PointMaze \textcolor{gray}{(4 datasets)}                 & 16.0                 & 38.2             & 27.8             & \underline{60.0} & 29.8          & \textbf{87.8}                       \\
                           &                               & AntMaze \textcolor{gray}{(4 datasets)}          & 19.8             & 31.8             & 35.0             & 53.0          & \underline{61.8} & \textbf{68.0}    \\
    \cmidrule{2-9}
                           & \multirow{3}{*}{Manipulation}
                           & Cube \textcolor{gray}{(4 datasets)}                      & 2.0                  & 22.5             & \underline{27.8} & 1.5              & 8.2           & \textbf{42.2}                       \\
                           &                               & Scene \textcolor{gray}{(1 dataset)}            & 5.0              & 42.0             & \underline{51.0} & 5.0           & 19.0             & \textbf{78.0}    \\
                           &                               & Puzzle \textcolor{gray}{(4 datasets)}           & 0.5              & 9.0              & \underline{36.8} & 0.2           & 2.0              & \textbf{51.0}    \\
    \cmidrule{2-9}
                           &                               & \cellcolor{gray!15}Overall \textcolor{gray}{(17 datasets)} & \cellcolor{gray!15}9.3 & \cellcolor{gray!15}26.4 & \cellcolor{gray!15}\underline{32.9} & \cellcolor{gray!15}27.3 & \cellcolor{gray!15}25.1 & \cellcolor{gray!15}\textbf{63.2} \\
    \midrule
    \multirow{5}{*}{Pixel} & \multirow{1}{*}{Locomotion}
                           & AntMaze \textcolor{gray}{(4 datasets)}                   & 5.0                  & 9.0              & 5.2              & 1.5              & \textbf{68.2} & \underline{61.0}                    \\
    \cmidrule{2-9}
                           & \multirow{3}{*}{Manipulation}
                           & Cube \textcolor{gray}{(4 datasets)}                      & 7.2                  & \underline{21.0} & 13.2             & 16.8             & 13.5          & \textbf{34.8}                       \\
                           &                               & Scene \textcolor{gray}{(1 dataset)}            & 12.0             & \underline{25.0} & 12.0             & 10.0          & 11.0             & \textbf{33.0}    \\
                           &                               & Puzzle \textcolor{gray}{(4 datasets)}           & 4.2              & \textbf{28.2}    & 6.5              & 0.2           & 4.8              & \underline{18.3} \\
    \cmidrule{2-9}
                           &                               & \cellcolor{gray!15}Overall \textcolor{gray}{(13 datasets)} & \cellcolor{gray!15}6.0 & \cellcolor{gray!15}19.8 & \cellcolor{gray!15}8.6 & \cellcolor{gray!15}6.5 & \cellcolor{gray!15}\underline{27.5} & \cellcolor{gray!15}\textbf{37.6} \\
    \bottomrule
  \end{tabular}
\end{table}

% Keep the heading and the wrapped figure on one page; a wrapfigure cut by a page break loses its
% caption.
% \Needspace{18\baselineskip}
\subsection{Q1: How Does \method{} Perform for Offline Goal Reaching?}
\label{sec:q-ogbench}

\textbf{Comparison against model-free baselines.}
Table~\ref{tab:summary} reports the results grouped by environment (full numerics in
Appendix~\ref{app:extended}). Averaged over the $17$ state datasets, \method{} nearly doubles the strongest baseline ($63.2$ vs.
$32.9$ for GCIQL) and leads every environment group. The gains
concentrate on manipulation and shrink on
navigation (AntMaze $68.0$ vs. $61.8$), where a reactive policy may already solve much of the task. This suggests that model-based planning can be more suitable for contact-rich tasks, where
test-time compute translates more directly to success. On pixel-based datasets, the gains become relatively modest ($37.6$ vs. $27.5$ for CRL) as \method{} trails behind in a long-horizon Visual AntMaze dataset.
\begin{wrapfigure}[14]{r}{0.296\textwidth}
  \centering
  \vspace{-0.4em}
  \includegraphics[width=\linewidth]{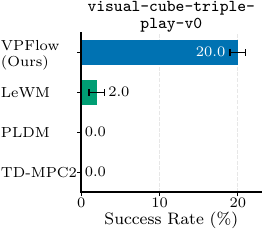}
  \caption{\textbf{Success rate of planners with latent world models} on
    \emph{the only} pixel dataset where the baselines do not completely fail.}
  \label{fig:latent-wm-baselines}
\end{wrapfigure}

\textbf{Comparison against latent planning methods.}
To our surprise, on 12 of the 13 pixel datasets, \emph{all three model-based baselines score zero},
so Figure~\ref{fig:latent-wm-baselines} reports the only dataset on which any one of them achieves a 
non-zero success rate. All three train their latent to predict a future state and then plan
over action sequences. On diverse play data, neither practice stands out as the optimal
design choice. We next reveal more about the possible failure modes on these two patterns.

% We attribute this partly to the unconstrained action space planning that may query the
% out-of-distribution planning objective.

\subsection{Q2: Value-Predictive or State-Predictive Latents?}
\label{sec:q-value-vs-state}

This experiment aims to verify which representations are more favorable for planning. The default model trains the rollout with the value-prediction
loss $\mathcal{L}_{\mathrm{VP}}$ (\Eqref{eq:value_prediction_loss}); the state-predictive model
trains it with the latent-prediction loss $\mathcal{L}_{\mathrm{LP}}$ (\Eqref{eq:state_loss}), which
matches the predicted latents to the encoded states. Other configurations stay the same. Figure~\ref{fig:value-state-comparison} tracks the success rate and the next-latent prediction error during training on four datasets.

\suppressfloats[t]
\begin{figure}[t]
  \centering
  \begin{subfigure}[T]{0.24\textwidth}
    \centering
    \includegraphics[width=\textwidth]{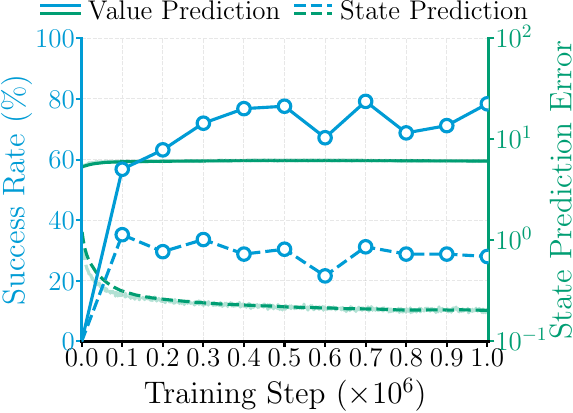}
    \caption{$\texttt{scene-play-v0}$}
    \label{fig:sub1}
  \end{subfigure}
  \hfill
  \begin{subfigure}[T]{0.24\textwidth}
    \centering
    \includegraphics[width=\textwidth]{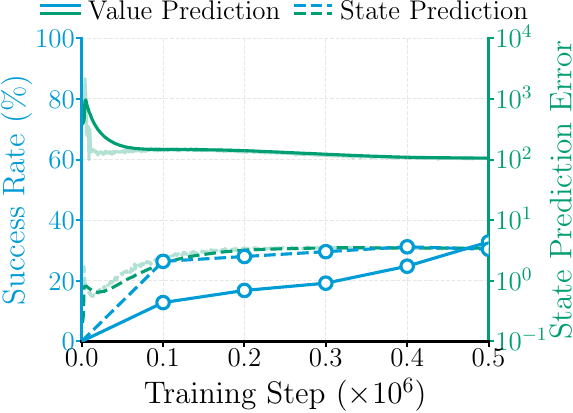}
    \caption{\texttt{visual-scene-play-v0}}
    \label{fig:sub2}
  \end{subfigure}
  \hfill
  \begin{subfigure}[T]{0.24\textwidth}
    \centering
    \includegraphics[width=\textwidth]{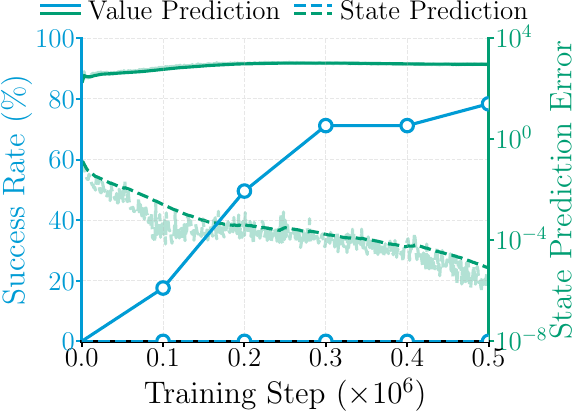}
    \caption{\texttt{visual-cube-single-}\\\texttt{play-v0}}
    \label{fig:sub3}
  \end{subfigure}
  \hfill
  \begin{subfigure}[T]{0.24\textwidth}
    \centering
    \includegraphics[width=\textwidth]{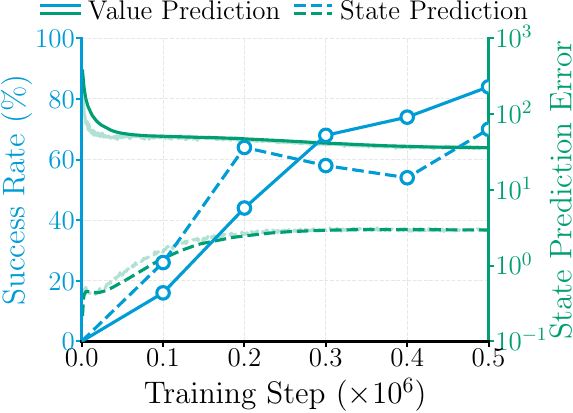}
    \caption{\texttt{visual-antmaze-large-}\\\texttt{navigate-v0}}
    \label{fig:sub4}
  \end{subfigure}
  \caption{\textbf{Comparison of value-predictive and state-predictive representations.} Success
    rate (left axis) and next-latent prediction error (right axis, log scale)
    during training, for value-predictive (solid) and state-predictive (dashed) models. The error
    is the mean Euclidean distance $\|\hat z_{t+h}-z_{t+h}\|_2$ between the predicted and encoded
    next latents. On state and pixel inputs, the value-predictive representation achieves higher
    success even when its state prediction error is larger by orders of magnitude.}
  \label{fig:value-state-comparison}
\end{figure}

\textbf{Lower state prediction error, worse planning.}
The state-predictive controller reduces the next-latent error by one to eight orders of magnitude,
yet its success does not improve on any of the four datasets and falls most sharply on
\texttt{scene-play} ($78.4\%$ to
$28.0\%$, Figure~\ref{fig:sub1}) and \texttt{visual-cube-single} ($79.2\%$ to $0.0\%$, Figure~\ref{fig:sub3}). A model that predicts its own next latent almost perfectly is therefore not
necessarily a model that plans well. Predicting the value is more essential for planning.

\looseness-1
\textbf{The failure mode is collapse, and value prediction prevents it.}
On \texttt{visual-cube-single} (Figure~\ref{fig:sub3}), the latent prediction term collapses to near
zero: the encoder finds a trivially predictable latent and the agent never reaches the goal, the
degeneracy that self-predictive objectives prevent through explicit anti-collapse techniques, none
of which our architecture has. In contrast, value prediction in \Eqref{eq:value_prediction_loss} promotes temporally ordered latents that implicitly prevent collapse. Appendix~\ref{app:explicit-regularization} provides ablations on explicit regularization over latent prediction.

\subsection{Q3: Does Steering Help, and Where Should It Search?}
\label{sec:q-planner}
\begin{figure}[htbp]
  \centering
  \begin{minipage}[htbp]{0.35\textwidth}
    \centering
    \scriptsize
    \setlength{\tabcolsep}{4pt}
    \captionof{table}{\textbf{Ablations on steering algorithm and search space.} Success rate
      ($\%$, $\uparrow$) averaged over the 12 locomotion (Loco.), 18 manipulation (Manip.), and all
      30 datasets. 
        Action-space
      planning
      worsens the success rate relative to direct action output.}
    \label{tab:planner-ablation}
    % \resizebox{\linewidth}{!}{
    \begin{tabular}{llccc}
      \toprule
      Planner      & Space  & Loco.         & Manip.        & All           \\
      \midrule
      w/o steering & --     & 57.5          & 8.3           & 28.0          \\
      w/ \acem{}   & action & 22.6          & 1.6           & 10.0          \\
      w/ \epscem{} & noise  & 53.2          & 31.5          & 40.2          \\
      \method{}    & noise  & \textbf{72.2} & \textbf{38.7} & \textbf{52.1} \\
      \bottomrule
    \end{tabular}
    % }
  \end{minipage}\hfill
  \begin{minipage}[htbp]{0.62\textwidth}
    \centering
    \includegraphics[width=\linewidth]{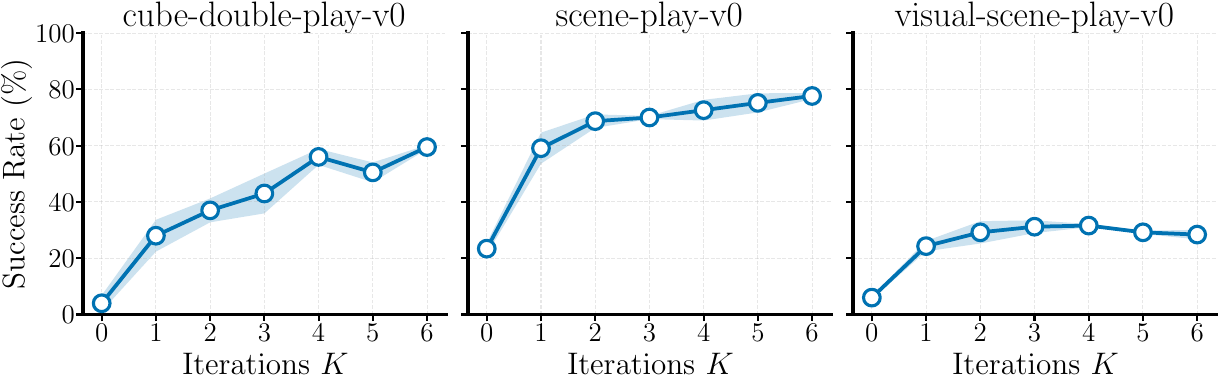}
    \caption{\textbf{Success rate ($\%$) against the number of steering iterations $K$} on
      \texttt{cube-double-play-v0}, \texttt{scene-play-v0}, and \texttt{visual-scene-play-v0}. More
      iterations generally steer the policy towards a higher success rate. $K=0$ means the reactive
      policy without steering.}
    \label{fig:steer-scaling}
  \end{minipage}
\end{figure}
\textbf{Test-time steering enables policy improvement with imagination.}
Without steering ($K{=}0$), \method{} executes the policy's actions directly and achieves inferior
performance (Table~\ref{tab:planner-ablation}). Extra test-time compute, by updating the noise
greedily towards maximal advantage in the imagined future, raises success as $K$ grows
(Figure~\ref{fig:steer-scaling}). This supports
\Eqref{eq:planning_objective} as an effective objective enabling policy improvement via search
without external signals, and planning with more compute can be made possible with imperfect world models.

\textbf{Noise space proves more effective than action space for planning.}
Replacing \epses{} with \epscem{} that updates noise with standard CEM lowers the success rate.
This is expected, as noise is updated by advantage on the decoded actions rather than the noise
itself. More notably, the success rate of CEM in action space, $\mathcal{A}$-CEM, is even worse than the direct output ($10.0$ vs. $28.0$) and not even close to its noise-space counterpart ($10.0$ vs. $40.2$). This can be attributed to the advantage exploitation of off-manifold
actions, and it explains why the action-space planning design choices in PLDM and LeWM
\citep{sobal2025pldm,maes2026leworldmodel} could be suboptimal with offline data. The
\emph{closed-loop}
planning in the noise space of a generative policy prevents this in that its decoded actions are
trained to be in the support of the data distribution.

\subsection{Q4: How Do the Latents Behave in Rollouts?}
\label{sec:q-diagnostics}

\begin{figure}[htbp]
  \centering
  \begin{subfigure}[t]{0.49\linewidth}
    \centering
    \includegraphics[width=\textwidth]{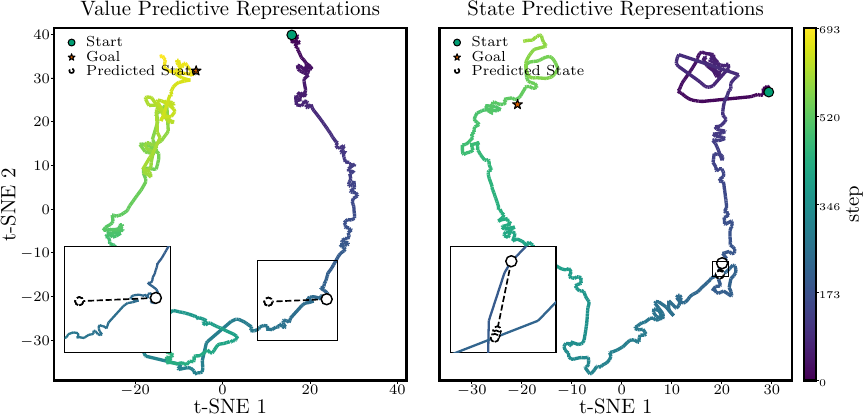}
    \caption{\texttt{visual-antmaze-large-navigate-v0} Task 1}
    \label{fig:latent-traj1}
  \end{subfigure}
  \begin{subfigure}[t]{0.49\linewidth}
    \centering
    \includegraphics[width=\textwidth]{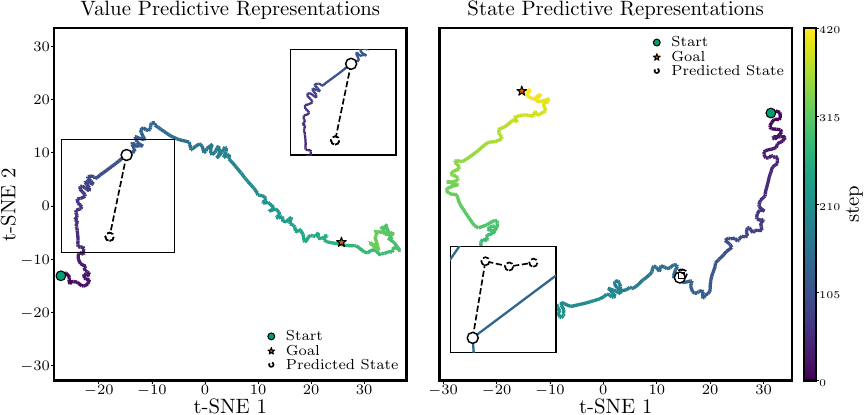}
    \caption{\texttt{visual-antmaze-large-navigate-v0} Task 3}
    \label{fig:latent-traj2}
  \end{subfigure}
  \caption{\textbf{t-SNE visualization of latent trajectories from successful Visual AntMaze episodes.} The solid circle is the encoded current state,
    and the dashed circles are the imagined latents via dynamics for that step. Under
    value-predictive training, the imagined latents leave the trajectory manifold at test time, but
    the agent can still reach the goal.}
  \label{fig:latent-traj}
\end{figure}

Figure~\ref{fig:latent-traj} depicts the geometry of imagined rollouts on Visual AntMaze to demonstrate the behavior of imperfect yet planning-effective learned world models.
Under value-predictive training, the dynamics model evolves latents to where the correct goal-reaching value is forecasted by the value function, without being constrained to match the corresponding embeddings from the encoder.
This helps explain why the value-predictive controller plans effectively despite the large next-latent prediction error demonstrated in Section~\ref{sec:q-value-vs-state}.
It also suggests how value prediction harnesses exploratory play data more effectively: closed-form value regression structures the latent space by temporal proximity to the goal, allowing it to leave out irrelevant routes and discover a latent geometry suited to reaching the goal.
In contrast, a state-predictive world model is trained to faithfully fit latent transitions in the dataset, which may limit its generalization to unseen goals.

\looseness-1
To examine this value-equivalent behavior more quantitatively, Figure~\ref{fig:rollout-level-sets} compares agents trained with value prediction $\mathcal{L}_{\mathrm{VP}}$ and latent prediction $\mathcal{L}_{\mathrm{LP}}$ on successful episodes from \texttt{visual-scene-play-v0}.
Starting from the same encoded state, each agent unrolls its own executed actions in both the real environment and the learned world model.
Although VP exhibits larger relative latent error across the 16-step rollout, its value gap is typically smaller than LP's (Figure~\ref{fig:rollout-consistency}).
The value field provides a microscopic illustration: while VP's observed and imagined latents are further apart in Euclidean space, they lie closer in the value space (Figure~\ref{fig:level-sets}).
The value-predictive latent also moves towards higher value, reflecting closer proximity to the goal.

\begin{figure}[htbp]
  \centering
  \begin{subfigure}[t]{0.49\linewidth}
    \centering
    \includegraphics[width=\textwidth]{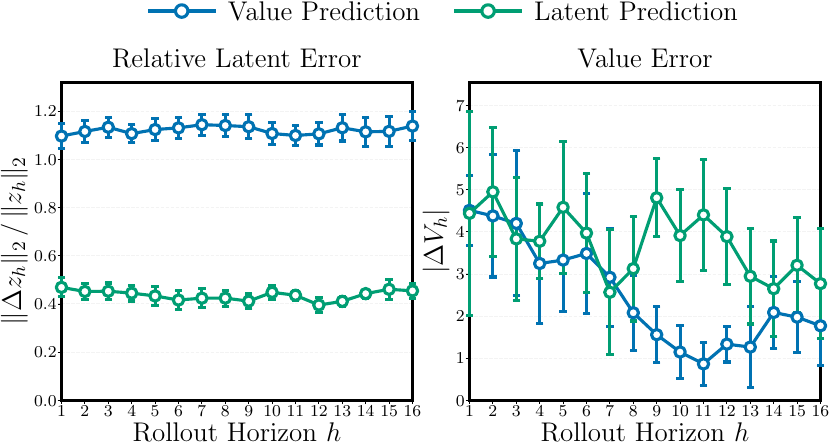}
    \caption{Relative latent error $\|\hat z_{h}-z_{h}\|_2/\|z_{h}\|_2$ and value error $|V(\hat z_{h}, z_g) - V(z_{h}, z_g)|$ along rollout horizons with mean and standard deviation over 4 evaluation seeds.}
    \label{fig:rollout-consistency}
  \end{subfigure}
  \hfill
  \begin{subfigure}[t]{0.49\linewidth}
    \centering
    \includegraphics[width=\textwidth]{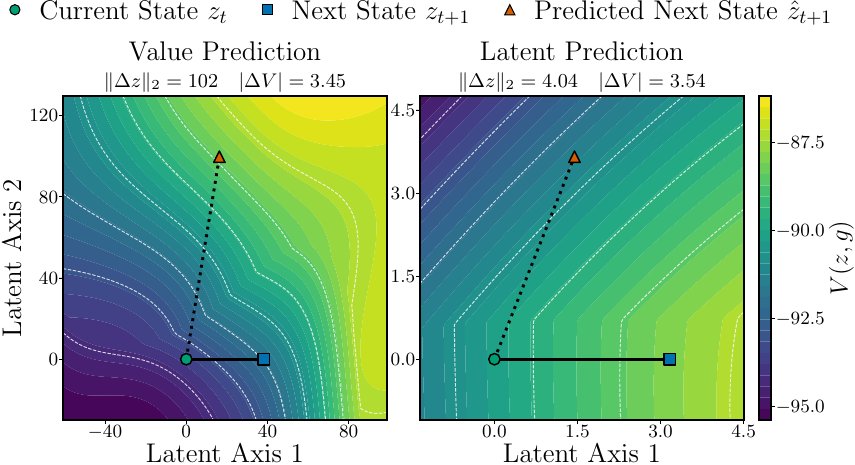}
    \caption{Value fields and level sets of each agent's learned critic measured on the current, observed, and predicted next latents.}
    \label{fig:level-sets}
  \end{subfigure}
  \looseness-1
  \caption{\textbf{Rollouts in the environment and learned world models, with their value fields.} With value prediction, the agent maintains a lower value error and reaches the goal successfully despite substantially larger latent prediction error.}
  \label{fig:rollout-level-sets}
\end{figure}
Taken together, these results demonstrate that accurate latent-state prediction is not strictly necessary for successful planning.
Imperfect world models can remain effective and even yield superior performance (Figure~\ref{fig:training-objective-ablation}), provided that they preserve value equivalence.
The evidence that the agent can successfully reach the goal despite substantial latent prediction error suggests that latent error is an unreliable indicator of planning success; instead, value prediction should serve as a more informative metric.
% The agent can reach the goal successfully despite the large latent prediction error, which implies that latent prediction error may not be a reliable indicator of planning performance and value prediction should be the more informative one.

\section{Conclusions}
\label{sec:conclusions}
\looseness-1
We have challenged the pervasive assumption that planning requires a state-predictive world model.
\method{} trains its observation encoder and latent dynamics solely via closed-form Monte Carlo
return regression---a supervision signal naturally available in sparse-reward offline goal
reaching---and plans at test time by steering a distilled one-step flow policy within its noise space to maximize imagined generalized advantage. 
Across 30 state and pixel datasets on OGBench, \method{} substantially improves upon model-free baselines and latent-predictive planners despite much higher next-latent prediction error.
In contrast, enforcing state prediction closes that error gap at the cost of degraded planning performance. 
Moreover, test-time noise steering reliably elevates the reactive policy prior, demonstrating that searching within an in-support noise space matters more than the specific optimization algorithm. More broadly, these findings indicate that effective latent planning favors value-predictive representations for world models and an in-distribution search space for planning, rather than a faithful state simulator. 
We discuss limitations and future directions in Appendix~\ref{app:discussion}.

%%%%%%%%%%%%%%%%%%%%%%%%%%%%%%%%%%%%%%%%%%%%%%%%%%%%%%%%%%%%
\clearpage
\end{maintextlayout}
% \newpage
% \onecolumn

\bibliographystyle{assets/plainnat}
\bibliography{ref}

%%%%%%%%%%%%%%%%%%%%%%%%%%%%%%%%%%%%%%%%%%%%%%%%%%%%%%%%%%%%
\clearpage
% \newpage
\beginappendix
% \appendix
\section{Extended Related Work}
\label{app:related}

This appendix expands upon the three lines of work discussed in Section~\ref{sec:related},
positioning \method{} within the broader literature.

\subsection{Offline and Goal-Conditioned Reinforcement Learning}

Offline RL learns policies from static datasets without environment interaction, mitigating
distribution shift either by biasing the policy towards the behavior distribution
\citep{fujimoto2019off,wu2020behavior,peng2019awr,fujimoto2021minimalist} or by regularizing critic
updates \citep{kumar2020conservative,kostrikov2022offline}. To capture multi-modal play data and
improve long-horizon credit assignment, Flow Q-learning (FQL) distills a conditional flow-matching
policy into a one-step network \citep{lipman2023flow,park2025flow}, and action chunking introduces
temporal abstraction \citep{li2025reinforcement,park2026horizon}. In goal-reaching settings
\citep{kaelbling1993learning}, universal value function approximators \citep{schaul2015universal}
and hindsight relabeling \citep{andrychowicz2017hindsight} extract supervision from reward-free
trajectories, which offline methods combine with iterated supervised learning
\citep{ghosh2021learning}, latent plans from play \citep{lynch2020learning}, contrastive learning
\citep{eysenbach2022contrastive}, quasi-metric representations \citep{wang2023qrl,myers2025quasimetric},
and subgoal hierarchies \citep{park2023hiql}, standardized by OGBench \citep{park2025ogbench}.
\method{} adopts a distilled action-chunk flow policy as its behavior prior
\citep{park2025flow,li2025reinforcement} but departs from these static, model-free methods in two
ways: it replaces TD, contrastive, or quasi-metric representation learning with closed-form
Monte Carlo return regression over multi-step imagined rollouts, and it scales test-time compute via
receding-horizon noise steering rather than acting greedily at deployment.

\subsection{Model-Based RL and World Models}

Model-based RL dates back to Dyna \citep{sutton1991dyna} and Gaussian-process policy search
\citep{deisenroth2011pilco}, with deep variants learning dynamics via pixel reconstruction
\citep{watter2015e2c,finn2017foresight,hafner2019learning} or latent imagination
\citep{ha2018world,hafner2020dream,hafner2025dreamerv3,hafner2025dreamerv4} and limiting compounding
error through short rollouts \citep{janner2019mbpo,park2026mac}. 
Offline model-based methods such as MOPO \citep{yu2020mopo}, MOReL \citep{kidambi2020morel}, COMBO
\citep{yu2021combo}, and MBOP \citep{argenson2021mbop} penalize or truncate rollouts where the model is uncertain, or plan with an ensemble model and a behavior prior like MBOP \citep{argenson2021mbop}.
Yet they still assume that accurate state prediction is a prerequisite for
planning. 
Value-equivalent models take an alternative position \citep{grimm2020value,m2023model}: value
prediction networks (VPN) \citep{oh2017vpn} and MuZero \citep{schrittwieser2020muzero} train
dynamics purely through predicted rewards, values, and policies. 
\method{} belongs to the value-equivalent family, but unlike online methods VPN and MuZero, it exploits closed-form hindsight Monte Carlo returns in offline goal reaching as the sole supervision for both the encoder and dynamics, eliminating target networks, EMA, state-matching loss, and TD learning from training.

\subsection{Predictive Representations}

Predictive representations characterize a state by its future outcomes rather than its sensory
appearance \citep{littman2001predictive,carvalho2024predictive}. One line of work builds on
successor representations and features
\citep{dayan1993improving,barreto2017successor,barreto2018transfer,borsa2019usfa}, forward-backward
representations \citep{touati2021fb,blier2021successor}, and behavior foundation models
\citep{pirotta2024bfm,park2024hilp} to predict discounted future occupancy for zero-shot reactive
readout. 
A parallel line learns with a self-predictive objective that forecasts future latent embeddings, relying on target encoders, stop-gradients \citep{schwarzer2021dataefficient,guo2022byolexplore,tang2023understanding,ni2024bridging,fujimoto2025mrq,bagatella2026tdjepa}, or explicit regularizers \citep{lecun2022path,dawid2024introduction,assran2025v,vanassel2025jepa} to prevent collapse. Whereas occupancy-based representations omit multi-step lookahead search at test time and self-predictive world models require explicit anti-collapse heuristics without exploiting temporal-distance signals, \method{} bridges both perspectives: it supervises multi-step latent rollouts with goal-conditioned values, which implicitly prevents collapse and organizes the latent space around temporal proximity to the goal.

\subsection{Planning in Latent Space}

Sampling-based optimizers such as CEM and MPPI \citep{rubinstein1999cem,williams2017mppi} are
standard planners in world models \citep{hafner2019learning,janner2019mbpo,pmlr-v162-hansen22a},
complemented in visual and offline domains by energy-based or functional-distance costs
\citep{du2019ebm,tian2021mbvp}, hierarchical subgoal search
\citep{li2022hierarchical,zhang2026hierarchical}, and generative adversarial behavior priors in
GOPlan \citep{wang2024goplan}. 
Closest to our setting, PLDM \citep{sobal2025pldm} and LeWorldModel \citep{maes2026leworldmodel} train latent-predictive dynamics with VICReg \citep{bardes2022vicreg}
or SIGReg \citep{balestriero2025lejepa} regularization and plan in action space via CEM to minimize
Euclidean distance to the goal embedding, whereas recent generative policy methods steer diffusion
or flow noise at test time using external rewards or online interaction
\citep{salimans2017evolution,yang2024diffusiones,wagenmaker2025dsrl,patil2026goldenticket}. In
contrast, \method{} avoids both the out-of-distribution action queries of open-loop action-space CEM
and the external reward requirements of prior noise-steering methods: it searches directly in the noise space of its distilled flow policy to maximize imagined generalized advantage under its value-predictive world model, keeping planned actions on the data manifold.

\section{Implementation Details}
\label{app:impl}

This section specifies the datasets (Section~\ref{app:data}), network
 details (Section~\ref{app:networks}), training objectives (Section~\ref{app:objectives}),
   optimization (Sections~\ref{app:train-loop} and \ref{app:optim}), evaluation protocol
   (Section~\ref{app:eval}), planner variants of \method{} (Section~\ref{app:planner-variants}),
   other model-based planning baselines in the latent space
   (Section~\ref{app:model-based-latent-planners}), and software and hardware
   (Section~\ref{app:compute}). Configurations and hyperparameters are detailed in
Appendix~\ref{app:hparams}.

\subsection{Datasets and Preprocessing}
\label{app:data}

\textbf{Datasets.}
We use the official OGBench datasets~\citep{park2025ogbench} covering both locomotion and robotic
manipulation tasks without
modification. The state suite comprises 17 datasets: \texttt{pointmaze-\{medium, large, giant,
  teleport\}-navigate}, \texttt{antmaze-\{medium, large, giant, teleport\}-navigate},
\texttt{cube-\{single, double, triple, quadruple\}-play}, \texttt{scene-play}, and
\texttt{puzzle-\{3x3, 4x4, 4x5, 4x6\}-play}. The pixel suite comprises the 13 \texttt{visual-}
counterparts of the AntMaze, Cube, Scene, and Puzzle datasets, with $64\times64\times3$
observations. Each dataset has five evaluation tasks specified by start-goal pairs.
% Visual-pointmaze is used only as a diagnostic environment and is excluded from all averages.

\textbf{Training samples and goal relabeling.}
A training sample is a length-$H$ segment $(s_t,a_t,\ldots,a_{t+H-1})$ drawn uniformly from the
dataset trajectories, with the chunk length $H$ in Table~\ref{tab:hparams-train}. The value goal
$g^V$ ($g$ in Section~\ref{sec:vpr}) is sampled from a mixture of three sources, with the
probabilities in Table~\ref{tab:hparams-train}: the current observation $s_t$; a future observation
$s_{t+d}$ on the same trajectory, with the offset $d$ drawn from the geometric distribution
$\mathrm{Geom}(1-\gamma)$; and a uniformly random dataset observation. For each offset
$h\in\{0,\ldots,H\}$, the MC target is $R_{t+h}=(1-\gamma^{d_h})/(\gamma-1)$ with $d_h$ as the
number of temporal steps from $s_{t+h}$ to $g^V$, and the mask $M_{t+h}$ is one if and only if $g^V$
is on the same trajectory at or after index $t+h$.
Random goals therefore contribute no value target.
The policy goal $g^\pi$ ($g$ in Section~\ref{sec:flow-policy}) is randomly sampled from a future
observation on the same trajectory.

\textbf{Data augmentation.}
For pixel-based tasks, we apply random-crop augmentation (edge padding, then cropping back to
$64\times64$; Table~\ref{tab:hparams-train}) using one shared crop per sample for the sampled
observation and both goals, except for the Visual AntMaze environment, following the practice in
OGBench.

\subsection{Networks}
\label{app:networks}

\textbf{Observation encoder.}
State-based tasks use the identity encoder $E_\phi(s)=s$. Pixel-based tasks use one IMPALA-Small
convolutional
encoder \citep{espeholt2018impala} shared by
observations and goals, followed by a multilayer perceptron (MLP) head that produces the latent
(Table~\ref{tab:hparams-arch}).

\begin{figure}[htbp]
  \centering
  \includegraphics[width=\textwidth]{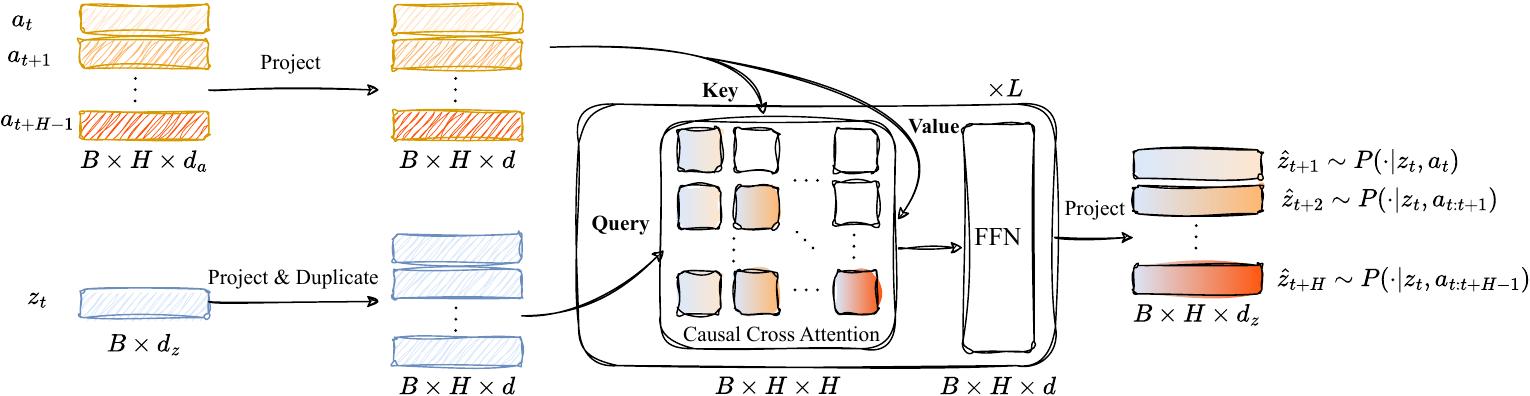}
  \caption{\textbf{World model architecture.} A Transformer with cross-attention and causal mask
    enables parallel rollouts of future latents induced by action chunks. $B$ is the batch size, $H$
    the chunk length, $d_a$ the action dimension, $d_z$ the latent dimension, $d$ the Transformer
    width.}
  \label{fig:world_model}
\end{figure}

\textbf{Dynamics model.}
The predictive world model $f_\phi$ maps the current latent $z_t$ and the action sequence
$a_{t:t+H-1}$ to subsequent latents $\hat{z}_{t+1:t+H}$.
To enable parallel rollouts, we instantiate $f_\phi$ as a Transformer \citep{vaswani2017attention}
with cross-attention and causal mask in which the current latent $z_t$ is the query and action
chunks $a_{t:t+H-1}$ are keys and values, as depicted in Figure~\ref{fig:world_model}.
It uses rotary position embeddings \citep{su2024roformer}, and its configuration is listed in
Table~\ref{tab:hparams-arch}.
The prediction for prefix $h$ attends only to $a_{t:t+h-1}$, so all $H$ prefix latents of a chunk are
produced in one parallel call.

\textbf{Value heads.}
Following previous practices to handle value overestimation and uncertainty
\citep{fujimoto2021minimalist,kostrikov2022offline}, we adopt a twin-value-head structure, where
each head is an MLP with layer normalization \citep{ba2016layernorm} and GELU activations
\citep{hendrycks2016gaussian} (Table~\ref{tab:hparams-arch}), acting on the
concatenation of the encoded state $z$ and goal $z_g$. Policy improvement and planning use the minimum
of the two heads, \ie, $\min\{V_{\phi,1},V_{\phi,2}\}$. The default MC objective uses no target
value network.

\textbf{Flow policy.}
The velocity field $v_\theta$ and the distilled one-step decoder $\pi_\theta$ are Transformers with
sinusoidal position embeddings whose size is tuned per dataset
(Table~\ref{tab:hparams-dataset}). Each network outputs action chunks of length
$H$, the same as the horizon of training samples.

State and goal conditioning adopts a similar design to that of $\pi_0$ \citep{black2024pi_0}: the
projected latent $z$ and goal $z_g$ are prefix tokens that the $H$ action tokens attend to, and the
flow time $u$ enters through a sinusoidal embedding fused into each action token by a two-layer MLP.
The prefix tokens attend only to each other while the action tokens attend to both the prefix and
themselves bidirectionally. The velocity field $v_\theta$ and the one-step policy $\pi_\theta$ share
this design.

\subsection{Objectives}
\label{app:objectives}

\textbf{Value-predictive model.}
\method{} trains its dynamics model and observation encoders by minimizing
$\mathcal{L}_{\mathrm{VP}}$ in \Eqref{eq:value_prediction_loss}.
The squared error is summed over the two value heads, and the per-sample loss is clipped at a
dataset-specific threshold.

\textbf{Conditional flow matching.}
Behavior cloning regularization is a critical technique to reduce OOD actions in offline RL
\citep{fujimoto2021minimalist}. Nonetheless, trajectories in the offline play dataset $\mathcal{D}$
are mostly multi-modal, so a uni-modal Gaussian policy is not expressive enough as the behavior
prior. We follow FQL \citep{park2025flow} in using conditional flow matching as the
behavior-cloning loss and a distilled one-step flow policy to predict multimodal actions.

Conditional flow matching \citep{lipman2023flow} learns a velocity field $v_\theta$ that
transports samples from a source distribution $\epsv\sim\mathcal{N}(\vzero,I)$ to a provided
action distribution $\va$ along the linear path $\vx_u=(1-u)\epsv+u\va$ with flow time
$u\in[0,1]$, whose target velocity is $\va-\epsv$:
\begin{equation}
  \mathcal{L}_{\mathrm{CFM}}(\theta)
  \coloneqq
  \mathbb{E}_{(s,\va,g)\sim\mathcal{D},\,\epsv,\,u}
  \!\left[
    \left\|v_\theta(\vx_u,u,E_\phi(s),E_\phi(g))-(\va-\epsv)\right\|_2^2
    \right].
  \label{eq:cfm}
\end{equation}

\textbf{Distillation.}
The distillation trains the one-step flow policy $\pi_\theta(z, z_g, \epsv)$, conditioned on
the encoded state $z$ and goal $z_g$, to match the action chunks decoded from the multi-step
integration of velocity $v_\theta$ using the same noise $\epsv\sim\mathcal{N}(\vzero,I)$ by
minimizing:
\begin{equation}
  \mathcal{L}_{\mathrm{Distill}}(\theta)
  \coloneqq
  \mathbb{E}_{(s,g)\sim\mathcal{D},\,\epsv}
  \Big[
    \big\|
    \pi_\theta(E_\phi(s),E_\phi(g),\epsv)
    -\operatorname{sg}\!\big(\operatorname{EulerIntegration}(v_\theta,\epsv)\big)
    \big\|_2^2
    \Big],
  \label{eq:distill_loss}
\end{equation}
where $\operatorname{sg}$ is the stop-gradient operation and
$\operatorname{EulerIntegration}(v_\theta,\epsv)$ is the action chunk obtained by integrating
$v_\theta$
with the Euler method from $u=0$ (the noise
$\epsv$) to $u=1$ with the number of steps in Table~\ref{tab:hparams-train}.

\textbf{Policy improvement.}
The policy minimizes \Eqref{eq:actor_loss} with the predictive model frozen.
Policy improvement imagines the future states with one chunk ($H$ actions) with the value term
in~\Eqref{eq:q_function} divided by its stop-gradient batch mean $|Q_\theta|$, stabilized by a small
constant $\epsilon_Q$.

\textbf{State-predictive model.}
\label{app:state-predictive-model}
The (latent) state-predictive model described in Section~\ref{sec:q-value-vs-state} trains the value
heads on encoded states $z_{t+h}=E_\phi(s_{t+h}), h\in[0, H]$ and includes an additional term with
the weight $\lambda_{\mathrm{state}}$ (Table~\ref{tab:hparams-train}) to train the rollout $\hat
z_{t+h}$ to predict the next latent $z_{t+h}$, using the latent-prediction (LP) loss:
\begin{equation}
      \mathcal{L}_{\mathrm{LP}}(\phi;\lambdastate)
    \coloneqq \mathbb{E}_{(\vs_t,\va_t, \vs_{t+1})\sim\mathcal{D},\,g\sim\mathcal{S}}\left[
    \frac{1}{H+1}\sum_{h=0}^{H}
    M_{t+h} \odot \|V_{\phi}(z_{t+h},z_{g})-R_{t+h}\|_2^2 +
\lambdastate\frac{1}{H}\sum_{h=1}^{H}\|\hat z_{t+h}-z_{t+h}\|_2^2\right].
    \label{eq:state_loss}
\end{equation}
The target latent $z_{t+h}$ has no stop-gradient operation, so the pixel encoder can be updated by
gradients from self-prediction.
% The next-latent prediction error reported in Figure~\ref{fig:value-state-comparison} is the batch
% mean of $\|\hat z_{t+h}-z_{t+h}\|_2$.

% Keep the wrapped algorithm and its heading on the same page.
% \Needspace{28\baselineskip}
\subsection{Training Loop}
\label{app:train-loop}
\begin{wrapfigure}[19]{r}{0.5\textwidth}
  \SetAlCapFnt{\footnotesize}
  \SetAlCapNameFnt{\footnotesize}
  \SetAlgoSkip{}
  \setlength{\algomargin}{1em}
\begin{algorithm}[H]
  \caption{\method{} training.}
  \label{alg:train}
  \KwIn{offline dataset $\mathcal{D}$; chunk length $H$; policy update interval $U$; training steps
    $T$;
    learning rates $\eta_\phi,\eta_\theta$}
  \KwOut{trained parameters $\phi$ and $\theta$}
  Initialize $\phi$ for $(E_\phi, f_\phi, V_{\phi,1}, V_{\phi,2})$ and $\theta$ for $(v_\theta,
    \pi_\theta)$\;
  \For{$k=1,\ldots,T$}{
    $(s_{t:t+H},\va_t=a_{t:t+H-1})\sim\mathcal{D}$; sample goals $g^V,g^\pi$\;
    $z_t\leftarrow E_\phi(s_t)$, $z_{g^V}\leftarrow E_\phi(g^V)$\;
    $\hat z_t\leftarrow z_t$\;
    $\hat z_{t+h}\leftarrow f_\phi(z_t,a_{t:t+h-1})$, $h=1,\ldots,H$
      \Comment*[r]{\Eqref{eq:latent_dynamics}}
    $d_h\leftarrow\operatorname{steps}(s_{t+h}\to g^V)$, $h=0,\ldots,H$ \Comment*[r]{Temporal distance to goal}
    $R_{t+h}\leftarrow(1-\gamma^{d_h})/(\gamma-1)$ \Comment*[r]{\Eqref{eq:mc_target}}
    $M_{t+h}\leftarrow\mathds{1}_{\{d_h\ge 0\}}$\;
    $\phi\leftarrow\operatorname{AdamStep}(\phi,\nabla_\phi\mathcal{L}_{\mathrm{VP}};\eta_\phi)$
    \Comment*[r]{\Eqref{eq:value_prediction_loss}, $\mathcal{L}_{\mathrm{VP}}$}
    \If{$k \bmod U = 0$}{
      % \Comment{Freeze $\phi$; differentiate through the rollout}
      $z_t\leftarrow\operatorname{sg}(E_\phi(s_t))$,
        $z_{g^\pi}\leftarrow\operatorname{sg}(E_\phi(g^\pi))$\;
      $\epsv\sim\mathcal{N}(\vzero,I)$\;
      $\va\leftarrow\operatorname{clip}(\pi_\theta(z_t,z_{g^\pi},\epsv),-1,1)$\;
      $\hat z_{t+H}\leftarrow f_\phi(z_t,\va)$\;
      $Q_\theta\leftarrow\gamma^H\min_{j\in\{1,2\}}V_{\phi,j}(\hat z_{t+H},z_{g^\pi})$\;
      $\theta\leftarrow\operatorname{AdamStep}(\theta,\nabla_\theta\mathcal{L}_\pi;\eta_\theta)$
      \Comment*[r]{\Eqref{eq:actor_loss}, $\mathcal{L}_{\pi}$}
    }
  }
  \KwRet $(\phi,\theta)$\;
\end{algorithm}
\end{wrapfigure}

Algorithm~\ref{alg:train} summarizes the training process in Figure~\ref{fig:vpflow-objective}.
The operator $\operatorname{AdamStep}$ denotes an Adam update step.
The goal-sampling rules and target masks follow Section~\ref{app:data}; $\operatorname{steps}$
returns
$-1$ for goals on another trajectory or before the predicted state.
Hyperparameters are specified in Table~\ref{tab:hparams-train}.

\subsection{Optimization}
\label{app:optim}

All networks are trained with Adam \citep{kingma2015adam}, with the
optimizer settings in Tables~\ref{tab:hparams-train} and~\ref{tab:hparams-dataset}. The number of
training steps follows the
benchmarking standards in OGBench \citep{park2025ogbench}. The predictive model with value heads and
the policy networks are updated alternately at every step, \ie, the policy update interval
$U=1$.

\subsection{Planning and Evaluation Protocol}
\label{app:eval}

\textbf{Planning setup.}
At each planning step, we encode the current observation and goal and maximize the imagined
advantage $A$ in \Eqref{eq:planning_objective}, keeping the model and policy weights fixed.
The default horizon is one training-sized action chunk ($n_c=1$, $H=16$).
We execute the first $\Hexec$ actions and replan from the resulting observation.
Shared planner hyperparameters like the population size $N$ and elite count $\Nelite$ are listed in
Table~\ref{tab:plan-shared}, and dataset-specific hyperparameters like execution horizons
$H_{\mathrm{exec}}$ are listed in Table~\ref{tab:plan-dataset}.

\textbf{Noise-space decoding.}
The distilled policy maps source noise to an action chunk, $\va=\pi_\theta(z,z_g,\epsv)$, with
$\epsv\sim\mathcal{N}(\vzero,I)$ in $\mathbb{R}^{Hd_a}$ during training, where $d_a$ is the action dimension.
Planning optimizes this noise through the decoder $\operatorname{Dec}$ in
Algorithm~\ref{alg:steering};
$\epsv=\vzero$ gives the deterministic policy output, which we call the policy mode.
For multiple chunks ($n_c\ge 1$), the decoder conditions each chunk $c$ on the latent reached by the
preceding chunk using its noise vector $\epsv_c$, where $\epsv=(\epsv_0,\ldots,\epsv_{n_c-1})$,
keeping dynamics queries within the training context length.
The score in \Eqref{eq:planning_objective} then extends over $n_cH$ steps.
Shorter planning horizons use a prefix of the decoded chunk.
Setting $K=0$ bypasses search and executes the policy mode.
\begin{figure}[htbp]
  \centering
  \begin{subfigure}[T]{0.71\textwidth}
    \centering
    \includegraphics[width=\linewidth]{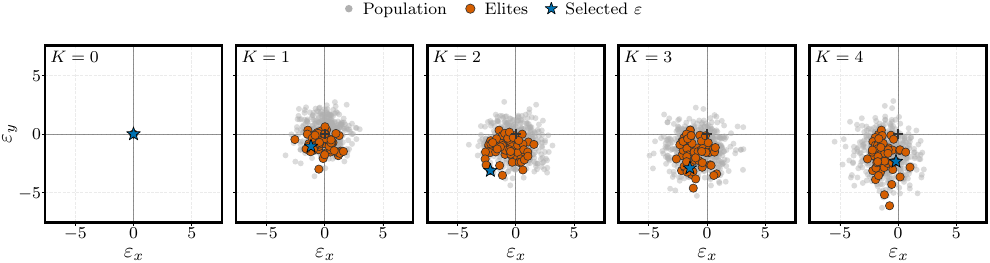}
    \caption{\textbf{Evolution of the noise population.}}
    \label{fig:es}
  \end{subfigure}\hfill
  \begin{subfigure}[T]{0.275\textwidth}
    \centering
    \includegraphics[width=\linewidth]{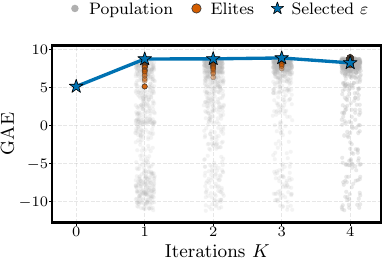}
    \caption{\textbf{GAE is optimized via noise-space search.}}
    \label{fig:es_gae}
  \end{subfigure}
  \caption{\textbf{Noise-space evolutionary search with \epses{} on
    \texttt{pointmaze-giant-navigate-v0}.}
    Gray points denote candidate noises, orange points denote elites, and the blue star marks the
      selected noise at each generation.
    The right panel shows the corresponding advantage scores.}
  \label{fig:es_visualization}
\end{figure}

\textbf{Zeroth-order steering.}
\epses{} (Algorithm~\ref{alg:steering}) uses forward evaluations of $A$, allowing search without
informative gradients.
It initializes $\Nelite$ elites at the policy mode and generates $N$ Gaussian mutations per
iteration, distributed evenly among the elites.
The mutation scale is annealed as $\sigma_k=\max(\sigma_{\min},\sigma_{\max}\rho^k)$ for
$k=0,\ldots,K-1$.
Each parent is replaced only by a child with a higher score, so every elite's score remains at least
as high as the initial policy score and the best score is non-decreasing.
The final plan is sampled from the elites with probabilities proportional to $\exp(\beta A_e)$,
retaining several high-scoring alternatives rather than selecting only the maximum.
The selected plan therefore also matches or exceeds the policy mode's score under the learned
objective $A$.
Figure~\ref{fig:es_visualization} illustrates the population and score evolution on PointMaze.

% Keep the wrapped algorithm beside the text that introduces first-order planning.
% \Needspace{26\baselineskip}
\begin{wrapfigure}[22]{r}{0.5\textwidth}
  \SetAlCapFnt{\footnotesize}
  \SetAlCapNameFnt{\footnotesize}
  \SetAlgoSkip{}
  \setlength{\algomargin}{1em}
\begin{algorithm}[H]
  \caption{\epssgd{} and \epsadam{}: first-order noise steering.}
  \label{alg:grad-steer}
  \KwIn{state $s_t$, goal $g$; optimizer $o\in\{\mathrm{SGD},\mathrm{Adam}\}$; steps $K\ge0$;
    step size $\eta$; decay $\rho$; execution length $\Hexec$;
    decoder $\operatorname{Dec}$ of Algorithm~\ref{alg:steering}}
  \KwOut{first $\Hexec$ actions of the final iterate}
  $z_t\leftarrow E_\phi(s_t)$, $z_g\leftarrow E_\phi(g)$\;
  \If{$K=0$}{
    \KwRet $\operatorname{Dec}(\vzero)[{:}\Hexec]$\;
  }
  $\epsv\leftarrow\vzero$ \Comment*[r]{Policy initialization at zero}
  $\vm\leftarrow\vzero$, $\vv\leftarrow\vzero$\;
  \For{$k=1,\ldots,K$}{
    $\eta_k\leftarrow\eta\rho^{k-1}$ \Comment*[r]{Annealed step size}
    $\vu\leftarrow\nabla_{\epsv}A(z_t,\operatorname{Dec}(\epsv),z_g)$
      \Comment*[r]{\Eqref{eq:grad_steer}}
    \eIf{$o=\mathrm{SGD}$}{
      $\epsv\leftarrow\epsv+\eta_k\vu$\;
    }{
      $\vm\leftarrow0.9\,\vm+0.1\,\vu$\;
      $\vv\leftarrow0.999\,\vv+0.001\,\vu^2$\;
      $\hat\vm\leftarrow\vm/(1-0.9^k)$, $\hat\vv\leftarrow\vv/(1-0.999^k)$\;
      $\epsv\leftarrow\epsv+\eta_k\hat\vm/(\sqrt{\hat\vv}+10^{-8})$\;
    }
  }
  \KwRet $\operatorname{Dec}(\epsv)[{:}\Hexec]$\;
\end{algorithm}
\end{wrapfigure}

\textbf{First-order steering.}
In practice, we find that first-order planners generally yield higher success rates than \epses{} on
locomotion tasks (Table~\ref{tab:steering}).
As a result, we steer the noise by gradient ascent with stochastic gradient descent (SGD) or Adam,
and denote the resulting planners \epssgd{} and \epsadam{}, respectively.
The main results in Table~\ref{tab:summary}, with complete evaluations in
Table~\ref{tab:state-pixel-main}, use these first-order planners for locomotion and state-based
\texttt{cube-single-play-v0} datasets, and \epses{} for the remaining manipulation datasets.

Both first-order planners (Algorithm~\ref{alg:grad-steer}) optimize a single noise candidate
initialized at $\epsv^{(0)}=\vzero$.
Each of the $K$ ascent steps differentiates through the policy decoder, latent dynamics, and value
function:
\begin{equation}
  \vu^{(k)}
  =
  \nabla_{\epsv} A\big(z_t,\operatorname{Dec}(\epsv),z_g\big)\Big|_{\epsv=\epsv^{(k-1)}}.
  \label{eq:grad_steer}
\end{equation}
\epssgd{} updates $\epsv^{(k)}=\epsv^{(k-1)}+\eta_k\vu^{(k)}$, with $\eta_k=\eta\rho^{k-1}$.
\epsadam{} uses bias-corrected Adam ascent \citep{kingma2015adam} with the same step-size schedule
and the moment constants in Table~\ref{tab:plan-shared}.
Adam rescales coordinates using running gradient moments, so its step size is tuned separately from
that of \epssgd{}.
Powers, square roots, and vector division are element-wise.
Noise is unconstrained, while decoded actions are clipped to $[-1,1]$.
We execute the final iterate without the elitist score guarantee of \epses{}.

\textbf{Planning complexity.}
Ignoring initialization and final decoding, \epses{} evaluates $KN$ candidate rollouts, whereas
first-order steering uses $K$ forward-backward evaluations of one candidate.
For a fixed model and rollout horizon, let $C_{\mathrm{fwd}}$ and $C_{\mathrm{bwd}}$ denote the
per-candidate complexity of decoding and scoring, and of backpropagation, respectively.
Under sequential computation, the total time complexities are $\mathcal{O}(KN C_{\mathrm{fwd}})$ for
\epses{} and $\mathcal{O}(K(C_{\mathrm{fwd}}+C_{\mathrm{bwd}}))$ for first-order steering.
The planner comparisons use separately tuned settings rather than matched compute budgets.

\textbf{Evaluation.}
We evaluate the final training checkpoint on each dataset's five official OGBench tasks, using $25$
episodes per task in vectorized environments \citep{park2025ogbench}.
Each episode retains its task goal, and success is determined by the environment's native indicator
within its time limit.
For each training seed, the dataset success rate is averaged over the five tasks.
We report the mean and standard deviation of success rate across 8 training seeds for state inputs
and 4 for pixel inputs, the same as in OGBench \citep{park2025ogbench}.

\subsection{Planner Variants}
\label{app:planner-variants}
The CEM variants in Table~\ref{tab:planner-ablation} optimize the same advantage $A$ as \epses{},
but refit a diagonal Gaussian at each iteration rather than retain and mutate individual elites.
Each iteration scores $N$ candidates and refits the mean and standard deviation using the top
$\Nelite$ candidates with normalized weights $w_e\propto\exp(\beta A_e)$.
This distribution refit does not preserve \epses{}'s monotonic elite-score guarantee.

\textbf{Noise-space CEM.}
\epscem{} (Algorithm~\ref{alg:eps-cem}) applies the CEM refitting scheme in the policy's noise
space, initializing the sampling distribution as $\mathcal{N}(\vzero,I)$.
Candidate noises are decoded through $\operatorname{Dec}$ of Algorithm~\ref{alg:steering} and ranked
by
$A$.
After each refit, the coordinate-wise standard deviations $\vsigma$ are clipped between $\sigma_{\min}$ and
the annealed upper bound $\sigma_k=\max(\sigma_{\min},\sigma_{\max}\rho^k)$, with $k$ indexed
from zero.
The final plan is decoded from an elite sampled with the same softmax weights.

\textbf{Action-space CEM.}
\acem{} (Algorithm~\ref{alg:a-cem}) searches directly over action sequences, warm-starting the
initial mean at the policy's direct output $\vmu_a=\operatorname{Dec}(\vzero)$ and setting the
initial
standard deviation to $\vsigma_a=\sigma_{\max}\vone$.
This initialization is recomputed at each planning step from the current observation, without
carrying over the previous plan.
Candidates are clipped to $[-1,1]$ before scoring, and the sampling standard deviations use the same
annealed bounds.
The executed plan is the final fitted action mean.

Both variants execute the first $\Hexec$ actions, return the direct policy output when $K=0$, and
add no execution noise.
They reuse \epses{}'s planning hyperparameters on each dataset (Tables~\ref{tab:plan-shared}
and~\ref{tab:plan-dataset}).
In the pseudocode below, $\operatorname{TopK}$ returns elite indices and $\vone$ is the all-ones
vector; vector squaring, square roots, and clipping are element-wise.

\begingroup
\SetAlCapFnt{\footnotesize}
\SetAlCapNameFnt{\footnotesize}
\SetAlgoSkip{}
\setlength{\algomargin}{1em}
\SetInd{0.35em}{0.65em}
\setlength{\parskip}{0pt}
\noindent
\begin{minipage}[t]{0.48\textwidth}
  \vspace{0pt}
  \begin{algorithm}[H]
    \caption{\epscem{}: CEM in noise space.}
    \label{alg:eps-cem}
    \KwIn{state $s_t$, goal $g$; population $N$; elites $1\le\Nelite\le N$; iterations $K\ge0$;
      scale bounds $\sigma_{\min},\sigma_{\max}$; decay $\rho$; inverse temperature $\beta$;
      execution length $\Hexec$; decoder $\operatorname{Dec}$ of Algorithm~\ref{alg:steering}}
    \KwOut{first $\Hexec$ actions of the selected plan}
    $z_t\leftarrow E_\phi(s_t)$, $z_g\leftarrow E_\phi(g)$\;
    \If{$K=0$}{
      \KwRet $\operatorname{Dec}(\vzero)[{:}\Hexec]$\;
    }
    $\vmu\leftarrow\vzero$, $\vsigma\leftarrow\vone$ \Comment*[r]{Gaussian noise initialization}
    \For{$k=0,\ldots,K-1$}{
      $\sigma_k\leftarrow\max(\sigma_{\min},\sigma_{\max}\rho^k)$\;
      \For{$i=1,\ldots,N$}{
        $\epsv^{(i)}\sim\mathcal{N}(\vmu,\operatorname{diag}(\vsigma^2))$ \Comment*[r]{Noise-space
          sampling}
        $A_i\leftarrow A(z_t,\operatorname{Dec}(\epsv^{(i)}),z_g)$
          \Comment*[r]{\Eqref{eq:planning_objective}}
      }
      $\mathcal{E}\leftarrow\operatorname{TopK}((A_i)_{i=1}^{N},\Nelite)$\;
      $w_i\leftarrow\exp(\beta A_i)/\sum_{j\in\mathcal{E}}\exp(\beta A_j)$,
        $i\in\mathcal{E}$\;
      $\vmu\leftarrow\sum_{i\in\mathcal{E}}w_i\epsv^{(i)}$\;
      $\vsigma\leftarrow\sqrt{\sum_{i\in\mathcal{E}}w_i(\epsv^{(i)}-\vmu)^2}$\;
      $\vsigma\leftarrow\operatorname{clip}(\vsigma,\sigma_{\min},\sigma_k)$\;
    }
    $i^\star\sim\operatorname{Categorical}((w_i)_{i\in\mathcal{E}})$\;
    \KwRet $\operatorname{Dec}(\epsv^{(i^\star)})[{:}\Hexec]$\;
  \end{algorithm}
\end{minipage}\hfill
\begin{minipage}[t]{0.48\textwidth}
  \vspace{0pt}
  \begin{algorithm}[H]
    \caption{\acem{}: CEM in action space.}
    \label{alg:a-cem}
    \KwIn{state $s_t$, goal $g$; population $N$; elites $1\le\Nelite\le N$; iterations $K\ge0$;
      scale bounds $\sigma_{\min},\sigma_{\max}$; decay $\rho$; inverse temperature $\beta$;
      execution length $\Hexec$; decoder $\operatorname{Dec}$ of Algorithm~\ref{alg:steering}}
    \KwOut{first $\Hexec$ actions of the fitted mean}
    $z_t\leftarrow E_\phi(s_t)$, $z_g\leftarrow E_\phi(g)$\;
    \If{$K=0$}{
      \KwRet $\operatorname{Dec}(\vzero)[{:}\Hexec]$\;
    }
    $\vmu_a\leftarrow\operatorname{Dec}(\vzero)$ \Comment*[r]{Action warm start}
    $\vsigma_a\leftarrow\sigma_{\max}\vone$\;
    \For{$k=0,\ldots,K-1$}{
      $\sigma_k\leftarrow\max(\sigma_{\min},\sigma_{\max}\rho^k)$\;
      $\vsigma_a\leftarrow\operatorname{clip}(\vsigma_a,\sigma_{\min},\sigma_k)$\;
      \For{$i=1,\ldots,N$}{
        $\va^{(i)}\sim\mathcal{N}(\vmu_a,\operatorname{diag}(\vsigma_a^2))$
          \Comment*[r]{Action-space
          sampling}
        $\va^{(i)}\leftarrow\operatorname{clip}(\va^{(i)},-1,1)$\;
        $A_i\leftarrow A(z_t,\va^{(i)},z_g)$
          \Comment*[r]{\Eqref{eq:planning_objective}}
      }
      $\mathcal{E}\leftarrow\operatorname{TopK}((A_i)_{i=1}^{N},\Nelite)$\;
      $w_i\leftarrow\exp(\beta A_i)/\sum_{j\in\mathcal{E}}\exp(\beta A_j)$,
        $i\in\mathcal{E}$\;
      $\vmu_a\leftarrow\sum_{i\in\mathcal{E}}w_i\va^{(i)}$\;
      $\vsigma_a\leftarrow\sqrt{\sum_{i\in\mathcal{E}}w_i(\va^{(i)}-\vmu_a)^2}$\;
      $\vsigma_a\leftarrow\operatorname{clip}(\vsigma_a,\sigma_{\min},\sigma_k)$\;
    }
    \KwRet $\vmu_a[{:}\Hexec]$\;
  \end{algorithm}
\end{minipage}
\par\endgroup

\subsection{Model-Based Latent Planners}
\label{app:model-based-latent-planners}

We benchmark LeWM \citep{maes2026leworldmodel}, PLDM \citep{sobal2025pldm}, and TD-MPC2
\citep{hansen2024td} on the 13 pixel-based datasets in Appendix~\ref{app:data}.
Our implementations are adapted from the official
LeWM\footnote{\url{https://github.com/lucas-maes/le-wm}} and stable-worldmodel
\citep{maes_lld2026swm}\footnote{\url{https://github.com/galilai-group/stable-worldmodel}}
repositories.
Each baseline learns a latent world model from the offline data and uses CEM
\citep{rubinstein1999cem} to plan towards a goal image.
Table~\ref{tab:baselines} summarizes the configurations; the objectives and OGBench adaptations are
described below.

\begin{table}[!t]
  \centering
  \renewcommand{\arraystretch}{1.05}
  \footnotesize
  \begin{threeparttable}
  \caption{\textbf{Configurations of latent planning baselines on pixel-based OGBench.}
    Merged cells indicate shared settings; -- denotes an absent component or an inapplicable
      setting.}
  \label{tab:baselines}
  \newcommand{\spanii}[1]{\multicolumn{2}{>{\centering\arraybackslash}p{\dimexpr0.54\linewidth-2\tabcolsep\relax}}{#1}}
  \newcommand{\spaniii}[1]{\multicolumn{3}{>{\centering\arraybackslash}p{\dimexpr0.81\linewidth-2\tabcolsep\relax}@{}}{#1}}
  \begin{tabularx}{\linewidth}{@{}>{\raggedright\arraybackslash}p{0.19\linewidth}*{3}{>{\centering\arraybackslash}X}@{}}
    \toprule
    \textbf{Setting} & \textbf{LeWM} & \textbf{PLDM} & \textbf{TD-MPC2}\tnote{b} \\
    \midrule
    \rowcolor{gray!8}[0pt][0pt]\multicolumn{4}{@{}l@{}}{\textbf{Architecture}} \\
    Image resolution & \spanii{$224\times224$} & $64\times64$ \\
    Context / frameskip & \spanii{$3$ frames / $5$ steps} & $1$ frame / $1$ step \\
    Encoder          & \spanii{ViT-Tiny, patch size $14$}
                     & $4$ conv.\ layers, $32$ channels; kernels $7,5,3,3$ \\
    Latent dimension & \spanii{$192$} & $128$, SimNorm group $8$ \\
    Dynamics         & \spanii{Causal Transformer with AdaLN-Zero; $6$ blocks, width $192$;
                       $16$ attention heads of dimension $64$; FFN width $2048$; dropout $0.1$}
                     & $3$-layer MLP, width $384$; SimNorm output \\
    Projection MLPs  & \spanii{Encoder and predictor: $2$ layers, hidden width $2048$, BatchNorm, GELU} & -- \\
    Control heads    & \spanii{--}
                     & Reward, $5$ Q-heads, policy: $3$-layer MLPs, width $384$; Q dropout $0.01$ \\
    \midrule
    \rowcolor{gray!8}[0pt][0pt]\multicolumn{4}{@{}l@{}}{\textbf{Training objectives}} \\
    Loss / regularization & SIGReg $0.09$; $1024$ projections, $17$ integration knots
                     & Variance $18$; covariance $12$; temporal variance $0.7$; alignment $0.2$; IDM $0$
                     & Consistency $20$; reward $0.1$; Q-value $0.1$; entropy $10^{-4}$ \\
    TD settings      & \spanii{--} & Discount $\gamma=0.99$; loss decay $\rho=0.5$; target EMA $\tau=0.01$ \\
    Reward / Q bins  & \spanii{--} & Two-hot: $101$ bins on $[-6,2]$ in symlog space \\
    \midrule
    \rowcolor{gray!8}[0pt][0pt]\multicolumn{4}{@{}l@{}}{\textbf{Optimization}} \\
    Update steps    & \spaniii{$500{,}000$\tnote{a}} \\
    Batch size       & \spanii{$128$} & $256$ \\
    Optimizer        & \spaniii{AdamW} \\
    Learning rate    & \spanii{$5\times10^{-5}$} & $3\times10^{-4}$; encoder $0.3\times$, policy $0.1\times$ \\
    Weight decay     & \spanii{0.001} & 0.01 \\
    LR schedule      & \spanii{Linear warmup, cosine decay} & Constant \\
    Gradient-norm clip & \spanii{$1$} & $20$ \\
    \midrule
    \rowcolor{gray!8}[0pt][0pt]\multicolumn{4}{@{}l@{}}{\textbf{Test-time planning}} \\
    CEM              & \spaniii{$300$ candidates, $30$ elites, $30$ iterations; variance scale $1$} \\
    Plan / execution & \spanii{$5$ chunks ($25$ environment steps)} & $5$ environment steps \\
    Cost minimized   & \spanii{Terminal squared goal distance $\|\hat z_{\mathrm{final}}-z_g\|_2^2$}
                     & Negative predicted return $-J$ (\Eqref{eq:tdmpc2-plan}) \\
    Q penalty        & \spanii{--} & Coefficient $0.5$ \\
    Action normalization & \spanii{Z-score} & None \\
    \bottomrule
  \end{tabularx}
  \begin{tablenotes}[flushleft]
    \footnotesize
    \item[a] Update steps refer to updates per component (encoder, policy, etc.).
    \item[b] TD-MPC2 shares its image encoder between state and goal; its MLP hidden layers use
      LayerNorm and Mish,
    and its policy is a tanh-squashed Gaussian.
  \end{tablenotes}
  \end{threeparttable}
\end{table}

\paragraph{LeWM} learns a latent world model from pixels end to end. A
ViT-Tiny encoder maps each frame $s_t$ to an embedding $z_t=E_\phi(s_t)$, and a causal Transformer
$f_\phi$ predicts the embedding $\hat z_{t+F}$ of the next frame from the three preceding frames.
Consecutive frames are $F=5$ environment steps apart (the frameskip), and the $F$ actions between
them form an action chunk $\va_t=a_{t:t+F-1}$, which conditions the Transformer through AdaLN-Zero.
The objective combines next-latent prediction error with SIGReg \citep{balestriero2025lejepa},
which encourages the embeddings to follow an isotropic Gaussian to prevent collapse:
\begin{equation}
  \mathcal{L}_{\mathrm{LeWM}}(\phi)
  \coloneqq
  \mathbb{E}_{\mathcal{D}}\Big[\big\|\hat z_{t+F}-z_{t+F}\big\|_2^2\Big]
  + \alpha_{\mathrm{SIG}}\,\mathrm{SIGReg}\big(\{E_\phi(s)\}_{s\in\mathcal{B}}\big),
  \label{eq:lewm}
\end{equation}
where $\mathcal{B}$ is a mini-batch of observations from $\mathcal{D}$, and $\alpha_{\mathrm{SIG}}$
is the SIGReg weight.
Gradients flow through both predictions and targets, without a target encoder, stop-gradient, or
EMA.
At test time, CEM minimizes the distance between the goal embedding and the
embedding predicted with the planned action chunks.

\paragraph{PLDM} shares the architecture and prediction loss of LeWM, but
prevents collapse with VICReg-style regularizers \citep{bardes2022vicreg}, which keep the embedding
dimensions spread and decorrelated, and with a temporal alignment term that pulls consecutive
embeddings together. It also includes an auxiliary inverse-dynamics objective that predicts the
intervening action chunk from $z_t$ and $z_{t+F}$:
\begin{equation}
  \mathcal{L}_{\mathrm{IDM}}(\phi)
  \coloneqq
  \mathbb{E}_{\mathcal{D}}\Big[\big\|\va_t-\mathrm{IDM}_\phi(z_t,z_{t+F})\big\|_2^2\Big].
  \label{eq:idm}
\end{equation}
We set the weight of this term to zero in our benchmark, following the stable-worldmodel
configuration.
PLDM plans with the same goal-embedding distance as LeWM.

\paragraph{TD-MPC2} learns a decoder-free latent world model together with reward and action-value
functions.
A CNN with SimNorm encodes each frame as $z_t=E_\phi(s_t)$; MLPs parameterize the dynamics $f_\phi$,
reward $\hat r_\phi$, five Q-heads $Q_{\phi,i}$, and policy $\pi_\theta$.
Training combines latent prediction consistency, reward prediction, and TD learning of the Q-heads.
Our goal-conditioned variant uses hindsight rewards $r(s,g)$ and goal embeddings $z_g=E_\phi(g)$, as
detailed below.
At test time, CEM maximizes the predicted return over $H=5$ steps with a conservative terminal
value:
\begin{equation}
  J(a_{t:t+H-1}; z_t,z_g)
  \coloneqq
  \sum_{h=0}^{H-1}\gamma^{h}\,\hat r_\phi(\hat z_{t+h},a_{t+h},z_g)
  + \gamma^{H}\,\big(\bar q-0.5\,|\bar q|\,\sigma_q\big),
  \label{eq:tdmpc2-plan}
\end{equation}
where $\hat z_t\coloneqq z_t$, $\hat z_{t+h+1}=f_\phi(\hat z_{t+h},a_{t+h})$, and $\bar q$ and
$\sigma_q$ are the mean and standard deviation over the five heads of
$Q_{\phi,i}\big(\hat z_{t+H},\pi_\theta(\hat z_{t+H},z_g),z_g\big)$. CEM is initialized with a
rollout of $\pi_\theta$, as in stable-worldmodel.

\textbf{Goal-conditioned TD-MPC2.}
To support OGBench's reward-free datasets and five evaluation goals per dataset, we add goal
conditioning and hindsight reward relabeling to stable-worldmodel's TD-MPC2.
A shared CNN encodes the current and goal images into $z_t$ and $z_g$, each normalized by SimNorm.
The reward head, Q-heads, and policy additionally receive $z_g$, while the dynamics $f_\phi$ remains
goal-independent.

Hindsight rewards $r(s,g)$ use OGBench's success criteria, computed from the joint positions and
button states stored in the dataset.
For Visual AntMaze, the reward is zero within the goal tolerance and $-1$ otherwise.
For Visual Cube, Puzzle, and Scene, it equals the number of satisfied subgoals minus their total.
These subgoals concern cube positions in Cube, button states in Puzzle, and cubes, buttons, the
drawer, and the window in Scene.
TD-MPC2 thus receives simulator-state reward supervision and partial subgoal credit, providing more
informative signals than \method{}'s sparse goal reward.

\textbf{Adaptations to OGBench.}
We use the shared CEM settings in Table~\ref{tab:baselines} and execute the full planned actions
before replanning.
The horizon spans five action chunks ($25$ environment steps) for LeWM and PLDM, and five
environment steps for TD-MPC2.
We align the data, training budget, and task evaluation with \method{} as follows:
\begin{itemize}[leftmargin=*,itemsep=1pt,parsep=0pt,topsep=2pt]
  \item \emph{Data.} We convert the official pixel datasets to HDF5 format, preserving observations,
    actions, and the simulator states needed for hindsight rewards.
    Training samples are trajectory subsequences with frames spaced by the baseline's frameskip; a
      random $90\%$ of these samples forms the training split.
  \item \emph{Training.} Each optimizer receives $500{,}000$ updates, matching \method{}'s training
    budget on pixel inputs.
  \item \emph{Evaluation.} We replace stable-worldmodel's dataset-based start-goal sampling with
    OGBench's five built-in tasks.
    Each episode is reset with its task identifier and retains the task's goal image throughout.
    Success follows the environment's native indicator within its time limit, as for \method{}.
\end{itemize}

\subsection{Software and Hardware}
\label{app:compute}
% \todo{add GPU model and per-run wall-clock hours (state and pixel).}
All experiments were conducted on NVIDIA A100 GPUs. The implementation is
in JAX \citep{jax2018github} and Equinox
\citep{kidger2021equinox}\footnote{\url{https://github.com/patrick-kidger/equinox}}.

% The two steering families differ structurally in cost: \epses{} evaluates $512K$ decoded rollouts
% per re-steering step, whereas the first-order variants evaluate $K$ gradients of a single
% candidate, so equal $K$ is not equal compute. (Planner-cost benchmarking dropped; restore if time
% allows.)

% Top-align the remaining appendix float pages.
\makeatletter
\setlength{\@fptop}{0pt}
\makeatother
% \Needspace{10\baselineskip}
\section{Hyperparameters}
\label{app:hparams}

\subsection{Training and Architecture}
\label{app:train-arch-hparams}
Tables~\ref{tab:hparams-train}, \ref{tab:hparams-dataset}, and~\ref{tab:hparams-arch} specify the shared training settings, dataset-specific settings, and network architectures, respectively.
Dashes denote settings that are not applicable.

\begin{table}[!htbp]
  \centering
  % \footnotesize
  \setlength{\tabcolsep}{0pt}
  % \setlength{\aboverulesep}{1.5pt}
  % \setlength{\belowrulesep}{1.5pt}
  % \captionsetup{font=footnotesize,skip=3pt}
  \renewcommand{\arraystretch}{0.98}
  \begin{threeparttable}
    \caption{\textbf{Shared training hyperparameters.} Entries spanning both columns apply to both observation modalities. }
    \label{tab:hparams-train}
    \begin{tabularx}{\linewidth}{@{}l>{\centering\arraybackslash}X>{\centering\arraybackslash}X@{}}
      \toprule
      \textbf{Quantity} & \textbf{State} & \textbf{Pixel} \\
      \midrule
      \rowcolor{gray!8}\multicolumn{3}{l}{\textbf{Optimization}} \\
      Training steps $T$ & $1{,}000{,}000$ & $500{,}000$ \\
      Batch size $B$ & \multicolumn{2}{c}{$256$} \\
      Optimizer & \multicolumn{2}{c}{Adam} \\
      Adam moment decay rates $(\beta_1,\beta_2)$ & \multicolumn{2}{c}{$(0.9,0.999)$} \\
      Adam numerical constant & \multicolumn{2}{c}{$10^{-8}$} \\
      Learning-rate schedule & \multicolumn{2}{c}{Cosine decay to zero} \\
      Gradient-norm clip & \multicolumn{2}{c}{$10$} \\
      Policy update interval $U$ & \multicolumn{2}{c}{$1$} \\
      Per-sample VP-loss clipping & \multicolumn{2}{c}{Dataset-specific\tnote{a}} \\
      \midrule
      \rowcolor{gray!8}\multicolumn{3}{l}{\textbf{Objectives}} \\
      Training horizon / action chunk length $H$ & \multicolumn{2}{c}{$16$} \\
      Teacher-flow Euler steps & \multicolumn{2}{c}{$10$} \\
      Policy normalization constant $\epsilon_Q$ & \multicolumn{2}{c}{$10^{-6}$} \\
      State-predictive ablation weight $\lambda_{\mathrm{state}}$ & \multicolumn{2}{c}{$1$\tnote{b}} \\
      \midrule
      \rowcolor{gray!8}\multicolumn{3}{l}{\textbf{Goal sampling and augmentation}} \\
      Value-goal mixture (current / future / random) & \multicolumn{2}{c}{$0.2\,/\,0.5\,/\,0.3$} \\
      Future value-goal offset $d$ & \multicolumn{2}{c}{$\mathrm{Geom}(1-\gamma)$} \\
      Policy-goal sampling & \multicolumn{2}{c}{Uniform future observation on the same trajectory} \\
      Image resolution & -- & $64\times64\times3$ \\
      Random-crop padding (pixels) & -- & $3$ \\
      \bottomrule
    \end{tabularx}
    \begin{tablenotes}[flushleft]
      \footnotesize
      \item[a] The per-sample VP loss is clipped at a dataset-specific threshold ranging from $10^4$ to $10^5$.
      \item[b] The default value-predictive model uses $\lambda_{\mathrm{state}}=0$; the state-predictive ablation in Section~\ref{sec:q-value-vs-state} uses $1$.
    \end{tablenotes}
  \end{threeparttable}
\end{table}

\begin{table}[htbp]
  \centering
  % \footnotesize
  \setlength{\tabcolsep}{0pt}
  \renewcommand{\arraystretch}{0.98}
  \begin{threeparttable}
    \caption{\textbf{Dataset-specific training hyperparameters.} $\eta_\phi=\eta_\theta$ is the initial learning rate, $\gamma$ is the discount, and $\alpha_{\mathrm{Distill}}$ and $\alpha_{\mathrm{CFM}}$ are the weights of the distillation and conditional flow matching losses.
    The flow policy network size is specified as width / layers / attention heads.}
    \label{tab:hparams-dataset}
    \begin{tabular*}{\linewidth}{@{\extracolsep{\fill}}llcccccc@{}}
      \toprule
      \textbf{Input} & \textbf{Dataset} & \textbf{$\eta_\phi=\eta_\theta$} & \textbf{$\gamma$} & \textbf{$\alpha_{\mathrm{Distill}}$} & \textbf{$\alpha_{\mathrm{CFM}}$} & \makecell{\textbf{Flow network}\\\textbf{size}} & \makecell{\textbf{Crop}\\\textbf{probability}} \\
      \midrule
      \multirow{17}{*}{State} & \texttt{pointmaze-medium-navigate-v0} & $2\times10^{-4}$ & $0.99$ & $200$ & $1$ & $192\,/\,3\,/\,3$ & -- \\
       & \texttt{pointmaze-large-navigate-v0} & $2\times10^{-4}$ & $0.99$ & $200$ & $1$ & $192\,/\,3\,/\,3$ & -- \\
       & \texttt{pointmaze-giant-navigate-v0} & $3\times10^{-4}$ & $0.995$ & $100$ & $1$ & $192\,/\,3\,/\,3$ & -- \\
       & \texttt{pointmaze-teleport-navigate-v0} & $2\times10^{-4}$ & $0.99$ & $200$ & $1$ & $192\,/\,3\,/\,3$ & -- \\
      \cmidrule(lr){2-8}
       & \texttt{antmaze-medium-navigate-v0} & $3\times10^{-4}$ & $0.99$ & $400$ & $1$ & $384\,/\,4\,/\,6$ & -- \\
       & \texttt{antmaze-large-navigate-v0} & $3\times10^{-4}$ & $0.99$ & $400$ & $1$ & $384\,/\,4\,/\,6$ & -- \\
       & \texttt{antmaze-giant-navigate-v0} & $2\times10^{-4}$ & $0.995$ & $50$ & $6$ & $192\,/\,3\,/\,3$ & -- \\
       & \texttt{antmaze-teleport-navigate-v0} & $5\times10^{-4}$ & $0.99$ & $500$ & $1$ & $192\,/\,3\,/\,3$ & -- \\
      \cmidrule(lr){2-8}
       & \texttt{cube-single-play-v0} & $3\times10^{-5}$ & $0.99$ & $200$ & $1$ & $192\,/\,3\,/\,3$ & -- \\
       & \texttt{cube-double-play-v0} & $3\times10^{-5}$ & $0.99$ & $200$ & $1$ & $192\,/\,3\,/\,3$ & -- \\
       & \texttt{cube-triple-play-v0} & $2\times10^{-4}$ & $0.99$ & $200$ & $1$ & $192\,/\,3\,/\,3$ & -- \\
       & \texttt{cube-quadruple-play-v0} & $2\times10^{-4}$ & $0.99$ & $200$ & $1$ & $192\,/\,3\,/\,3$ & -- \\
      \cmidrule(lr){2-8}
       & \texttt{scene-play-v0} & $2\times10^{-4}$ & $0.99$ & $200$ & $1$ & $192\,/\,3\,/\,3$ & -- \\
      \cmidrule(lr){2-8}
       & \texttt{puzzle-3x3-play-v0} & $3\times10^{-4}$ & $0.99$ & $200$ & $1$ & $192\,/\,3\,/\,3$ & -- \\
       & \texttt{puzzle-4x4-play-v0} & $2\times10^{-4}$ & $0.99$ & $200$ & $1$ & $192\,/\,3\,/\,3$ & -- \\
       & \texttt{puzzle-4x5-play-v0} & $2\times10^{-4}$ & $0.99$ & $200$ & $1$ & $192\,/\,3\,/\,3$ & -- \\
       & \texttt{puzzle-4x6-play-v0} & $2\times10^{-4}$ & $0.99$ & $50$ & $1$ & $192\,/\,3\,/\,3$ & -- \\
      \midrule
      \multirow{13}{*}{Pixel} & \texttt{visual-antmaze-medium-navigate-v0} & $1\times10^{-4}$ & $0.99$ & $400$ & $1$ & $384\,/\,4\,/\,6$ & 0 \\
       & \texttt{visual-antmaze-large-navigate-v0} & $1\times10^{-4}$ & $0.99$ & $400$ & $1$ & $512\,/\,4\,/\,8$ & 0 \\
       & \texttt{visual-antmaze-giant-navigate-v0} & $1\times10^{-4}$ & $0.99$ & $400$ & $1$ & $512\,/\,4\,/\,8$ & 0 \\
       & \texttt{visual-antmaze-teleport-navigate-v0} & $1\times10^{-4}$ & $0.99$ & $400$ & $1$ & $384\,/\,4\,/\,6$ & 0 \\
      \cmidrule(lr){2-8}
       & \texttt{visual-cube-single-play-v0} & $3\times10^{-4}$ & $0.99$ & $400$ & $1$ & $384\,/\,4\,/\,6$ & 0.5 \\
       & \texttt{visual-cube-double-play-v0} & $1\times10^{-4}$ & $0.99$ & $400$ & $1$ & $384\,/\,4\,/\,6$ & 0.5 \\
       & \texttt{visual-cube-triple-play-v0} & $1\times10^{-4}$ & $0.99$ & $400$ & $1$ & $384\,/\,4\,/\,6$ & 0.5 \\
       & \texttt{visual-cube-quadruple-play-v0} & $1\times10^{-4}$ & $0.99$ & $400$ & $1$ & $384\,/\,4\,/\,6$ & 0.5 \\
      \cmidrule(lr){2-8}
       & \texttt{visual-scene-play-v0} & $2\times10^{-4}$ & $0.99$ & $200$ & $1$ & $192\,/\,3\,/\,3$ & 0.5 \\
      \cmidrule(lr){2-8}
       & \texttt{visual-puzzle-3x3-play-v0} & $1\times10^{-4}$ & $0.99$ & $400$ & $1$ & $192\,/\,3\,/\,3$ & 0.5 \\
       & \texttt{visual-puzzle-4x4-play-v0} & $5\times10^{-4}$ & $0.99$ & $800$ & $1$ & $384\,/\,4\,/\,6$ & 0.5 \\
       & \texttt{visual-puzzle-4x5-play-v0} & $3\times10^{-4}$ & $0.99$ & $400$ & $1$ & $192\,/\,3\,/\,3$ & 0.5 \\
       & \texttt{visual-puzzle-4x6-play-v0} & $2\times10^{-4}$ & $0.99$ & $100$ & $1$ & $384\,/\,4\,/\,6$ & 0.5 \\
      \bottomrule
    \end{tabular*}
  \end{threeparttable}
\end{table}

\begin{table}[htbp]
  \centering
  % \small
  \setlength{\tabcolsep}{0pt}
  \renewcommand{\arraystretch}{1.08}
  \begin{threeparttable}
    \caption{\textbf{Network architectures.} Entries spanning both columns are shared across modalities. $d$, $L$, and $d_z$ denote width, layer count, and latent dimension, as in Figure~\ref{fig:world_model}.}
    \label{tab:hparams-arch}
    \begin{tabularx}{\linewidth}{@{}l>{\centering\arraybackslash}X>{\centering\arraybackslash}X@{}}
      \toprule
      \textbf{Quantity} & \textbf{State} & \textbf{Pixel} \\
      \midrule
      \rowcolor{gray!8}\multicolumn{3}{l}{\textbf{Observation encoder $E_\phi$}} \\
      Architecture & Identity & IMPALA-Small\tnote{a} \\
      Convolutional stack channels & -- & $16\,/\,32\,/\,32$ \\
      Residual blocks per stack & -- & $1$ \\
      Convolution kernel / stride & -- & $3\times3$ / $1$ \\
      Max-pooling kernel / stride & -- & $3\times3$ / $2$ \\
      Output head & -- & Linear, $512$ units, GELU \\
      Latent dimension $d_z$ & Observation dimension & $512$ \\
      \midrule
      \rowcolor{gray!8}\multicolumn{3}{l}{\textbf{Latent dynamics $f_\phi$}} \\
      Architecture & \multicolumn{2}{c}{Transformer, cross-attention, causal mask} \\
      Width $d$ / layers $L$ / heads & \multicolumn{2}{c}{$192\,/\,3\,/\,3$} \\
      Feedforward multiplier & \multicolumn{2}{c}{$4$} \\
      Activation / normalization & \multicolumn{2}{c}{GELU / layer normalization} \\
      Dropout & \multicolumn{2}{c}{$0.1$} \\
      Position embedding & \multicolumn{2}{c}{Rotary} \\
      \midrule
      \rowcolor{gray!8}\multicolumn{3}{l}{\textbf{Value heads $V_{\phi}$}} \\
      Number of heads & \multicolumn{2}{c}{$2$} \\
      Hidden layers / output dimension & \multicolumn{2}{c}{$512\,/\,512\,/\,512$ / $1$} \\
      Activation / normalization & \multicolumn{2}{c}{GELU / layer normalization} \\
      \midrule
      \rowcolor{gray!8}\multicolumn{3}{l}{\textbf{Flow networks $v_\theta,\pi_\theta$}} \\
      Architecture & \multicolumn{2}{c}{Transformer, state and goal as prefix tokens} \\
      Width / layers / heads & \multicolumn{2}{c}{Dataset-specific (Table~\ref{tab:hparams-dataset})} \\
      Feedforward multiplier & \multicolumn{2}{c}{$4$} \\
      Activation / normalization & \multicolumn{2}{c}{GELU / layer normalization} \\
      Position embedding & \multicolumn{2}{c}{Sinusoidal} \\
      Flow-time embedding ($v_\theta$ only) & \multicolumn{2}{c}{Sinusoidal, fused via a two-layer MLP} \\
      Action-token attention & \multicolumn{2}{c}{Bidirectional\tnote{b}} \\
      \bottomrule
    \end{tabularx}
    \begin{tablenotes}[flushleft]
      \footnotesize
      \item[a] The pixel encoder shares its parameters between observations and goals.
      Convolutional residual blocks use ReLU; the output head uses GELU.
      \item[b] Prefix tokens attend only to the state and goal tokens; action tokens attend to both prefix tokens and the full action chunk.
    \end{tablenotes}
  \end{threeparttable}
\end{table}

\subsection{Planning}
\label{app:steer-hparams}

% \textbf{Hyperparameter search.}
We tuned planning hyperparameters separately for each dataset and planner using the candidate
values in Table~\ref{tab:plan-grid}.
Each configuration was evaluated on a fixed trained checkpoint, with goal-reaching success rate
as the selection criterion.
For each configuration, we varied the number of refinement iterations $K$ and selected the
smallest positive value achieving its highest observed success rate.
The number of action chunks $n_c$ determines the imagined rollout length.
Each chunk contains $H=16$ actions, so $n_c$ complete chunks span $n_cH$ environment steps.
The execution length $\Hexec$ determines how many actions are executed before replanning from
a new observation.
The $K=0$ baseline executes the direct policy output without planning.

Table~\ref{tab:plan-shared} presents the shared planning hyperparameters, and
Table~\ref{tab:plan-dataset} gives the settings for each dataset and planner.
Table~\ref{tab:steering} motivates the planner choices in Table~\ref{tab:plan-dataset}: we use first-order planning on locomotion tasks and state-based \texttt{cube-single-play-v0}, where it generally achieves higher success rates, and \epses{} on the remaining manipulation datasets.

\begin{table}[htbp]
  \centering
  \footnotesize
  \renewcommand{\arraystretch}{1.04}
  \begin{minipage}[t]{0.6\linewidth}
    \vspace{0pt}
    \caption{\textbf{Candidate values for planning hyperparameter search.}}
    \label{tab:plan-grid}
    % \begin{tabularx}{\linewidth}{@{}>{\raggedright\arraybackslash}p{0.38\linewidth}>{\raggedright\arraybackslash}X@{}}
    \begin{tabular*}{\linewidth}{@{\extracolsep{\fill}}ll@{}}
      \toprule
      \textbf{Hyperparameter} & \textbf{Candidate values} \\
      \midrule
      Iterations $K$ & $\{1,\ldots,10\}$ \\
      Maximum mutation scale $\sigma_{\max}$ & $\{0.1,~0.2,~0.3,~0.5,~0.8,~1.0\}$ \\
      Minimum mutation scale $\sigma_{\min}$ & $\{0,~0.05,~0.1\}$ \\
      Inverse temperature $\beta$ & $\{0.1,~0.3,~0.5,~0.7,~0.9\}$ \\
      Per-iteration decay $\rho$ & $\{0.5,~0.6,~0.7,~0.8,~0.9,~1.0\}$ \\
      Execution length $\Hexec$ & $\{1,~2,~3,~5,~8,~16\}$ \\
      Trace-decay coefficient $\lambda$ & $\{0,~0.1,~0.3,~0.5,~0.6,~0.7,~0.8,~0.9,~1.0\}$ \\
      SGD step size $\eta$ & $\{0.001,~0.01,~0.02,~0.03,~0.05,~0.1,~0.2\}$ \\
      Adam step size $\eta$ & $\{0.01,~0.02,~0.03,~0.05,~0.06,~0.1\}$ \\
      Population size $N$ & $\{300,~512,~1024\}$ \\
      Number of elites $\Nelite$ & $\{30,~64,~100\}$ \\
      Number of action chunks $n_c$ & $\{1,~2,~3\}$ \\
      \bottomrule
    \end{tabular*}
  \end{minipage}\hfill
      \begin{minipage}[t]{0.38\linewidth}
    \vspace{0pt}
    \caption{\textbf{Shared planning hyperparameters.} Dataset-specific planner settings are listed in Table~\ref{tab:plan-dataset}.}
    \label{tab:plan-shared}
    % \begin{tabularx}{\linewidth}{@{}>{\raggedright\arraybackslash}p{0.70\linewidth}>{\centering\arraybackslash}X@{}}
    \begin{tabularx}{\linewidth}{@{}>{\raggedright\arraybackslash}p{0.70\linewidth}>{\raggedright\arraybackslash}X@{}}
      \toprule
      \textbf{Quantity} & \textbf{Value} \\
      \midrule
      Action chunk length $H$ & $16$ \\
      Number of action chunks $n_c$ & $1$ \\
      Population size $N$ & $512$ \\
      Number of elites $\Nelite$ & $64$ \\
      Children per elite $\lceil N/\Nelite\rceil$ & $8$ \\
      Inverse temperature $\beta$ & $0.1$ \\
      Decoded action clipping range & $[-1,1]$ \\
      Adam moment decay rates $(\beta_1,\beta_2)$ & $(0.9,0.999)$ \\
      Adam numerical constant & $10^{-8}$ \\
      \bottomrule
    \end{tabularx}
  \end{minipage}
  % \par
  \bigskip
  \setlength{\tabcolsep}{0.0pt}
  \renewcommand{\arraystretch}{1.04}
  \captionsetup{position=top}
    \caption{\textbf{Dataset-specific planning hyperparameters.} $K$ is the number of planning iterations. $\sigma_{\max}$ and $\sigma_{\min}$ are the initial and minimum mutation scales; $\eta$ is the initial step size;
      $\rho$ is the per-iteration decay; $\lambda$ is the trace-decay coefficient in GAE;
      $\Hexec$ is the number of actions executed before replanning. The last column identifies the planner type used for the reported results: FO denotes the listed first-order planner, and ES denotes \epses{}.}
    \label{tab:plan-dataset}
    % \begin{tabular*}{\linewidth}{@{\extracolsep{\fill}}llcccccc cccccc c@{}}
    \begin{tabularx}{\linewidth}{@{\extracolsep{\fill}}llcccccc cccccc c@{}}
      \toprule
      & & \multicolumn{6}{c}{\epses{}} & \multicolumn{6}{c}{First-order} & \\
      \cmidrule(lr){3-8}\cmidrule(lr){9-14}
       \textbf{Input} & \textbf{Dataset} & $K$ & $\sigma_{\max}$ & $\sigma_{\min}$ & $\rho$ & $\lambda$ & $\Hexec$ & \textbf{Opt.} & $K$ & $\eta$ & $\rho$ & $\lambda$ & $\Hexec$ & \textbf{Planner} \\
      \midrule
      \multirow{17}{*}{State} & \texttt{pointmaze-medium-navigate-v0} & 1 & 0.5 & 0 & 0.5 & 1 & 16 & \epssgd{} & 1 & 0.05 & 1 & 1 & 16 & FO \\
       & \texttt{pointmaze-large-navigate-v0} & 1 & 0.1 & 0 & 0.5 & 1 & 16 & \epssgd{} & 6 & 0.05 & 1 & 1 & 16 & FO \\
       & \texttt{pointmaze-giant-navigate-v0} & 4 & 1 & 0.05 & 1 & 0.6 & 16 & \epssgd{} & 5 & 0.1 & 1 & 1 & 16 & FO \\
       & \texttt{pointmaze-teleport-navigate-v0} & 2 & 0.5 & 0 & 0.5 & 1 & 3 & \epssgd{} & 2 & 0.01 & 1 & 1 & 3 & FO \\
      \cmidrule(lr){2-15}
       & \texttt{antmaze-medium-navigate-v0} & 3 & 0.5 & 0 & 0.5 & 1 & 3 & \epssgd{} & 6 & 0.05 & 1 & 1 & 3 & FO \\
       & \texttt{antmaze-large-navigate-v0} & 2 & 0.1 & 0.05 & 1 & 1 & 3 & \epssgd{} & 5 & 0.2 & 1 & 1 & 3 & FO \\
       & \texttt{antmaze-giant-navigate-v0} & 7 & 0.2 & 0.05 & 1 & 1 & 3 & \epssgd{} & 4 & 0.05 & 1 & 1 & 3 & FO \\
       & \texttt{antmaze-teleport-navigate-v0} & 1 & 1 & 0 & 1 & 1 & 3 & \epssgd{} & 5 & 0.1 & 1 & 1 & 3 & FO \\
      \cmidrule(lr){2-15}
       & \texttt{cube-single-play-v0} & 3 & 0.5 & 0 & 1 & 1 & 3 & \epsadam{} & 7 & 0.05 & 1 & 1 & 3 & FO \\
       & \texttt{cube-double-play-v0} & 6 & 0.3 & 0 & 1 & 1 & 3 & \epsadam{} & 8 & 0.05 & 1 & 1 & 3 & ES \\
       & \texttt{cube-triple-play-v0} & 3 & 0.5 & 0 & 1 & 1 & 3 & \epsadam{} & 6 & 0.1 & 1 & 1 & 3 & ES \\
       & \texttt{cube-quadruple-play-v0} & 1 & 0.5 & 0 & 1 & 1 & 3 & \epsadam{} & 1 & 0.05 & 1 & 1 & 3 & ES \\
      \cmidrule(lr){2-15}
       & \texttt{scene-play-v0} & 5 & 0.5 & 0 & 1 & 1 & 3 & \epsadam{} & 6 & 0.05 & 1 & 1 & 3 & ES \\
      \cmidrule(lr){2-15}
       & \texttt{puzzle-3x3-play-v0} & 2 & 1 & 0 & 1 & 1 & 3 & \epsadam{} & 7 & 0.05 & 1 & 1 & 3 & ES \\
       & \texttt{puzzle-4x4-play-v0} & 2 & 1 & 0.1 & 1 & 1 & 1 & \epsadam{} & 7 & 0.1 & 1 & 1 & 3 & ES \\
       & \texttt{puzzle-4x5-play-v0} & 4 & 0.5 & 0 & 1 & 1 & 3 & \epsadam{} & 4 & 0.06 & 1 & 1 & 3 & ES \\
       & \texttt{puzzle-4x6-play-v0} & 5 & 1 & 0.1 & 1 & 1 & 3 & \epsadam{} & 6 & 0.1 & 1 & 1 & 3 & ES \\
      \midrule
      \multirow{13}{*}{Pixel} & \texttt{visual-antmaze-medium-navigate-v0} & 2 & 0.5 & 0 & 1 & 1 & 3 & \epsadam{} & 5 & 0.05 & 1 & 1 & 3 & FO \\
       & \texttt{visual-antmaze-large-navigate-v0} & 4 & 0.5 & 0.05 & 0.5 & 1 & 3 & \epsadam{} & 2 & 0.1 & 0.5 & 1 & 1 & FO \\
       & \texttt{visual-antmaze-giant-navigate-v0} & 3 & 0.5 & 0 & 1 & 1 & 8 & \epsadam{} & 4 & 0.05 & 0.5 & 1 & 8 & FO \\
       & \texttt{visual-antmaze-teleport-navigate-v0} & 2 & 0.5 & 0 & 1 & 1 & 3 & \epsadam{} & 5 & 0.03 & 1 & 1 & 3 & FO \\
      \cmidrule(lr){2-15}
       & \texttt{visual-cube-single-play-v0} & 2 & 0.5 & 0 & 1 & 1 & 3 & \epsadam{} & 6 & 0.1 & 0.6 & 1 & 3 & ES \\
       & \texttt{visual-cube-double-play-v0} & 1 & 0.5 & 0 & 1 & 1 & 3 & \epsadam{} & 3 & 0.1 & 1 & 1 & 3 & ES \\
       & \texttt{visual-cube-triple-play-v0} & 2 & 0.5 & 0.05 & 0.5 & 1 & 3 & \epsadam{} & 1 & 0.1 & 0.5 & 1 & 3 & ES \\
       & \texttt{visual-cube-quadruple-play-v0} & 1 & 0.5 & 0.05 & 0.5 & 1 & 3 & \epsadam{} & 4 & 0.1 & 0.7 & 1 & 3 & ES \\
      \cmidrule(lr){2-15}
       & \texttt{visual-scene-play-v0} & 3 & 0.5 & 0 & 1 & 1 & 3 & \epsadam{} & 4 & 0.1 & 0.6 & 1 & 3 & ES \\
      \cmidrule(lr){2-15}
       & \texttt{visual-puzzle-3x3-play-v0} & 4 & 0.5 & 0 & 1 & 1 & 3 & \epsadam{} & 4 & 0.1 & 0.6 & 1 & 3 & ES \\
       & \texttt{visual-puzzle-4x4-play-v0} & 5 & 0.5 & 0 & 1 & 1 & 8 & \epsadam{} & 1 & 0.05 & 1 & 1 & 3 & ES \\
       & \texttt{visual-puzzle-4x5-play-v0} & 2 & 1 & 0.05 & 1 & 1 & 3 & \epsadam{} & 1 & 0.03 & 1 & 1 & 3 & ES \\
       & \texttt{visual-puzzle-4x6-play-v0} & 4 & 1 & 0.05 & 1 & 1 & 3 & \epsadam{} & 1 & 0.03 & 1 & 1 & 3 & ES \\
      \bottomrule
    \end{tabularx}
\end{table}

% \Needspace{10\baselineskip}
\clearpage
\section{More Ablations on Key Design Choices}
\label{app:ablations}
We examine the planning objective used to score imagined trajectories and the effect of explicit regularization on latent-predictive training.

\subsection{Planning Objectives}
\label{app:gae}
\begin{figure}[htbp]
  \centering
  \captionsetup{width=\linewidth}
  \includegraphics[width=0.9\linewidth]{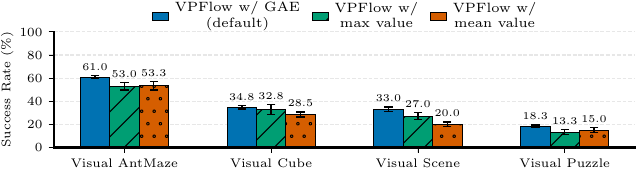}
  \caption{\textbf{Planning objective ablation.}
    Success rate (\%) is averaged within each pixel-based environment group.
    The objectives use the same trained checkpoint and planning hyperparameters for each dataset.
    Error bars denote $\pm$ one pooled within-group standard deviation.}
  \label{fig:planning-objective-ablation}
\end{figure}
GAE instantiates the MPC objective in \Eqref{eq:mpc} by using temporal-difference (TD) errors as potential-based shaped rewards.
Sparse goal-conditioned rewards provide no ranking signal among candidate trajectories that do not reach the goal within the planning horizon.
The TD error in \Eqref{eq:reshaped_reward}, $\delta_{t+h}=r_{t+h}+\gamma V_{t+h+1}-V_{t+h}$, provides a densified reward for credit assignment in planning, where $V_{t+h}\coloneqq V_\phi(\hat z_{t+h},z_g)$.
Maximizing \Eqref{eq:planning_objective} therefore corresponds to minimizing running costs $-(\gamma\lambda)^h\delta_{t+h}$ under the learned dynamics, with zero explicit terminal cost.

Potential-based reward shaping connects this dense objective to discounted-return maximization.
For $\gamma<1$ and a bounded potential, the infinite-horizon discounted shaped return differs from the original return only by the initial potential, preserving optimal policies \citep{ng1999policy}.
At the finite planning horizon, setting $\lambda=1$ gives the exact identity
\begin{equation}
  A_t\big|_{\lambda=1}
  = \underbrace{\sum_{h=0}^{H-1}\gamma^h r_{t+h}+\gamma^H V_{t+H}}_{H\text{-step return}}-V_t.
  \label{eq:gae-telescope}
\end{equation}
Summing the discounted TD errors therefore recovers the $H$-step MPC return minus the fixed initial value.
This identity holds for arbitrary value estimates, although planning performance depends on model and value accuracy.

An $H$-step return is consequently equivalent to GAE for planning at $\lambda=1$.
Adding the fixed baseline $V_t$ changes neither candidate rankings nor action gradients, so we omit this redundant ablation.
More generally, $A_t+V_t$ is the finite-horizon $\lambda$-return, which combines returns over different horizons and can change candidate rankings when $\lambda < 1$ \citep{schulman2016gae}.
Our implementation assigns a constant reward $r_{t+h}=-1$ to each imagined step without assuming any goal is reached.
For a fixed planning horizon and $\lambda=1$, the accumulated reward is identical across candidates, so maximizing GAE is equivalent to maximizing the predicted terminal value.

The ablations instead test whether directly aggregating intermediate values improves on this return-based score.
One alternative, the mean-value objective, is defined as the weighted average of predicted values along the rollout:
\begin{equation}
  C_{\text{mean}}(\va_t;z_t,z_g)
  \coloneq \frac{\sum_{h=1}^{H}\gamma^{h-1}V_{t+h}}{\sum_{h=1}^{H}\gamma^{h-1}}.
  \label{eq:mean-value-planning}
\end{equation}
This objective favors trajectories with high predicted values throughout the planning horizon, with earlier predictions weighted more heavily.
Unlike GAE at $\lambda=1$, it can rank a trajectory with higher intermediate values above one with a higher terminal value.

Max-value scoring instead takes the highest predicted value along the rollout:
\begin{equation}
  C_{\text{max}}(\va_t;z_t,z_g)
  \coloneq \max_{1\leq h\leq H} V_{t+h}.
  \label{eq:max-value-planning}
\end{equation}
This objective tests whether reaching a high-value state at any point within the planning horizon suffices for effective receding-horizon control.
It disregards subsequent decreases in predicted value but can be sensitive to value overestimation.

As shown in Figure~\ref{fig:planning-objective-ablation}, GAE achieved the highest mean success rate in all four visual environment groups.
Its success rates were $7.7$, $2.0$, $6.0$, and $3.3$ percentage points higher than the better alternative on Visual AntMaze, Cube, Scene, and Puzzle, respectively.
GAE is therefore the preferred planning objective among the evaluated alternatives.
Its equivalence to the $H$-step return at $\lambda=1$ justifies this empirical choice as the principled objective of MPC.

% \Needspace{8\baselineskip}
\subsection{Latent Prediction with Explicit Regularization}
\label{app:explicit-regularization}

\begin{figure}[htbp]
  \centering
  \captionsetup{width=\linewidth}
  \includegraphics[width=0.9\linewidth]{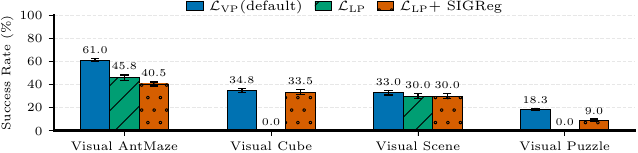}
  \caption{\textbf{Training objective ablation with explicit regularization.}
    Success rate (\%) is averaged within each visual environment group.
    Error bars denote $\pm$ one pooled within-group standard deviation across training seeds. Value prediction implicitly regularizes the representation through closed-form goal-conditioned value supervision and can be more effective than explicit regularization with SIGReg for planning.}
  \label{fig:training-objective-ablation}
\end{figure}

% \looseness-1
This ablation tests whether explicit representation regularization closes the planning-performance gap between latent-predictive and value-predictive training reported in Section~\ref{sec:q-value-vs-state}.
We compare our value-predictive objective $\mathcal{L}_{\mathrm{VP}}$ in \Eqref{eq:value_prediction_loss} with the latent-predictive objective $\mathcal{L}_{\mathrm{LP}}$ in \Eqref{eq:state_loss}, and its regularized variant
\begin{equation}
  \mathcal{L}_{\mathrm{LP+SIG}}(\phi)
  \coloneq \mathcal{L}_{\mathrm{LP}}(\phi;\lambdastate)
    + \alpha_{\mathrm{SIG}}\,\mathrm{SIGReg}\big(\{E_\phi(s)\}_{s\in\mathcal{B}}\big),
  \label{eq:lp-sigreg}
\end{equation}
where $\mathcal{B}$ is a mini-batch of dataset observations and $\alpha_{\mathrm{SIG}}$ controls the regularization strength and has the same value as in LeWM \citep{maes2026leworldmodel}.
SIGReg encourages an isotropic Gaussian embedding distribution to address representation collapse \citep{balestriero2025lejepa}.
All three objectives retain closed-form Monte Carlo value targets and share the architecture, policy-training objective, and training and planning hyperparameters. 

As shown in Figure~\ref{fig:training-objective-ablation}, latent prediction severely degraded the representations in the manipulation environments Visual Cube and Puzzle. While explicit regularization with SIGReg improved the latent-predictive representations, it did not close the gap to value-predictive training.
These results corroborate the benefits of value-predictive representations in planning through closed-form goal-conditioned value supervision over the widely used next-latent prediction, even with explicit representation regularization.

\clearpage
\section{Per-Dataset OGBench Results}
\label{app:extended}

Table~\ref{tab:state-pixel-main} reports the complete per-dataset results with state and pixel observations, summarized in Table~\ref{tab:summary}. We omit the hierarchical baseline: hierarchical implicit
Q-learning (HIQL) \citep{park2023hiql}, since \method{} is a flat controller.
Table~\ref{tab:steering} complements the aggregate planning comparison in Section~\ref{sec:q-planner} and Table~\ref{tab:planner-ablation} with per-dataset results for \epses{}, first-order noise steering, and the reactive policy ($K=0$).

\begin{table}[htbp]
  \centering
  \scriptsize
  \setlength{\tabcolsep}{3pt}
  \caption{\textbf{Complete success rates (\%) on OGBench state-based and pixel-based datasets.}
    Results are reported with mean $\pm$ standard deviation across 8 seeds for state input and 4 seeds for pixel input. Baseline numbers are
      from
    OGBench. The best result in each row is in bold. }
  \label{tab:state-pixel-main}
  \resizebox{\linewidth}{!}{
    \begin{tabular}{lll l *{5}{c} c}
      \toprule
      \textbf{Input} & \textbf{Env.} & \textbf{Type}                         & \textbf{Dataset}
                    & GCBC                                  & GCIVL                                   & GCIQL       & QRL                  & CRL                  & \method{}                                                 \\
      \midrule
      \multirow{17}{*}{State} & \multirow{4}{*}{PointMaze}
                    & \multirow{4}{*}{Nav.}
                    & \texttt{pointmaze-medium-navigate-v0} & $9_{\pm6}$                              & $63_{\pm6}$ & $53_{\pm8}$          & $82_{\pm8}$          & $29_{\pm7}$ & $\mathbf{98_{\pm2}}$                        \\
                    &&                                       & \texttt{pointmaze-large-navigate-v0}    & $29_{\pm6}$ & $45_{\pm5}$          & $34_{\pm3}$          & $86_{\pm9}$ & $39_{\pm7}$          & $\mathbf{99_{\pm1}}$ \\
                    &&                                       & \texttt{pointmaze-giant-navigate-v0}    & $1_{\pm2}$  & $0_{\pm0}$           & $0_{\pm0}$           & $68_{\pm7}$ & $27_{\pm10}$         & $\mathbf{90_{\pm3}}$ \\
                    &&                                       & \texttt{pointmaze-teleport-navigate-v0} & $25_{\pm3}$ & $45_{\pm3}$          & $24_{\pm7}$          & $4_{\pm4}$  & $24_{\pm6}$          & $\mathbf{64_{\pm2}}$ \\
      \cmidrule{2-10}
      &\multirow{4}{*}{AntMaze}
                    & \multirow{4}{*}{Nav.}
                    & \texttt{antmaze-medium-navigate-v0}   & $29_{\pm4}$                             & $72_{\pm8}$ & $71_{\pm4}$          & $88_{\pm3}$          & $95_{\pm1}$ & $\mathbf{96_{\pm1}}$                        \\
                    &&                                       & \texttt{antmaze-large-navigate-v0}      & $24_{\pm2}$ & $16_{\pm5}$          & $34_{\pm4}$          & $75_{\pm6}$ & $83_{\pm4}$          & $\mathbf{85_{\pm2}}$ \\
                    &&                                       & \texttt{antmaze-giant-navigate-v0}      & $0_{\pm0}$  & $0_{\pm0}$           & $0_{\pm0}$           & $14_{\pm3}$ & $16_{\pm3}$          & $\mathbf{33_{\pm3}}$ \\
                    &&                                       & \texttt{antmaze-teleport-navigate-v0}   & $26_{\pm3}$ & $39_{\pm3}$          & $35_{\pm5}$          & $35_{\pm5}$ & $53_{\pm2}$          & $\mathbf{58_{\pm3}}$ \\
      \cmidrule{2-10}
      &\multirow{4}{*}{Cube}
                    & \multirow{4}{*}{Play}
                    & \texttt{cube-single-play-v0}          & $6_{\pm2}$                              & $53_{\pm4}$ & $68_{\pm6}$          & $5_{\pm1}$           & $19_{\pm2}$ & $\mathbf{98_{\pm1}}$                        \\
                    &&                                       & \texttt{cube-double-play-v0}            & $1_{\pm1}$  & $36_{\pm3}$          & $40_{\pm5}$          & $1_{\pm0}$  & $10_{\pm2}$          & $\mathbf{60_{\pm1}}$ \\
                    &&                                       & \texttt{cube-triple-play-v0}            & $1_{\pm1}$  & $1_{\pm0}$           & $3_{\pm1}$           & $0_{\pm0}$  & $4_{\pm1}$           & $\mathbf{10_{\pm2}}$ \\
                    &&                                       & \texttt{cube-quadruple-play-v0}         & $0_{\pm0}$  & $0_{\pm0}$           & $0_{\pm0}$           & $0_{\pm0}$  & $0_{\pm0}$           & $\mathbf{1_{\pm0}}$  \\
      \cmidrule{2-10}
      &Scene         & Play                                  & \texttt{scene-play-v0}                  & $5_{\pm1}$  & $42_{\pm4}$          & $51_{\pm4}$          & $5_{\pm1}$  & $19_{\pm2}$          & $\mathbf{78_{\pm2}}$ \\
      \cmidrule{2-10}
      &\multirow{4}{*}{Puzzle}
                    & \multirow{4}{*}{Play}
                    & \texttt{puzzle-3x3-play-v0}           & $2_{\pm0}$                              & $6_{\pm1}$  & $\mathbf{95_{\pm1}}$ & $1_{\pm0}$           & $3_{\pm1}$  & $93_{\pm1}$                                 \\
                    &&                                       & \texttt{puzzle-4x4-play-v0}             & $0_{\pm0}$  & $13_{\pm2}$          & $26_{\pm3}$          & $0_{\pm0}$  & $0_{\pm0}$           & $\mathbf{79_{\pm4}}$ \\
                    &&                                       & \texttt{puzzle-4x5-play-v0}             & $0_{\pm0}$  & $7_{\pm1}$           & $14_{\pm1}$          & $0_{\pm0}$  & $1_{\pm0}$           & $\mathbf{20_{\pm2}}$ \\
                    &&                                       & \texttt{puzzle-4x6-play-v0}             & $0_{\pm0}$  & $10_{\pm2}$          & $\mathbf{12_{\pm1}}$ & $0_{\pm0}$  & $4_{\pm1}$           & $\mathbf{12_{\pm1}}$ \\
      \midrule
    \multirow{13}{*}{Pixel} & \multirow{4}{*}{AntMaze}
                    & \multirow{4}{*}{Nav.}
                    & \texttt{visual-antmaze-medium-navigate-v0} & $11_{\pm2}$                                  & $22_{\pm2}$          & $11_{\pm1}$          & $0_{\pm0}$   & $94_{\pm1}$  & $\mathbf{96_{\pm1}}$                        \\
                    &&                                            & \texttt{visual-antmaze-large-navigate-v0}    & $4_{\pm0}$           & $5_{\pm1}$           & $4_{\pm1}$   & $0_{\pm0}$   & $\mathbf{84_{\pm1}}$ & $\mathbf{84_{\pm1}}$ \\
                    &&                                            & \texttt{visual-antmaze-giant-navigate-v0}    & $0_{\pm0}$           & $1_{\pm1}$           & $0_{\pm0}$   & $0_{\pm0}$   & $\mathbf{47_{\pm2}}$ & $10_{\pm1}$          \\
                    &&                                            & \texttt{visual-antmaze-teleport-navigate-v0} & $5_{\pm1}$           & $8_{\pm1}$           & $6_{\pm1}$   & $6_{\pm3}$   & $48_{\pm2}$          & $\mathbf{54_{\pm2}}$ \\
      \cmidrule{2-10}
      &\multirow{4}{*}{Cube}
                    & \multirow{4}{*}{Play}
                    & \texttt{visual-cube-single-play-v0}        & $5_{\pm1}$                                   & $60_{\pm5}$          & $30_{\pm5}$          & $41_{\pm15}$ & $31_{\pm15}$ & $\mathbf{83_{\pm3}}$                        \\
                    &&                                            & \texttt{visual-cube-double-play-v0}          & $1_{\pm1}$           & $10_{\pm2}$          & $1_{\pm1}$   & $5_{\pm0}$   & $2_{\pm1}$           & $\mathbf{30_{\pm1}}$ \\
                    &&                                            & \texttt{visual-cube-triple-play-v0}          & $15_{\pm2}$          & $14_{\pm2}$          & $15_{\pm1}$  & $16_{\pm1}$  & $17_{\pm2}$          & $\mathbf{20_{\pm1}}$ \\
                    &&                                            & \texttt{visual-cube-quadruple-play-v0}       & $\mathbf{8_{\pm1}}$  & $0_{\pm0}$           & $7_{\pm1}$   & $5_{\pm1}$   & $4_{\pm1}$           & $6_{\pm0}$           \\
      \cmidrule{2-10}
      &Scene         & Play                                       & \texttt{visual-scene-play-v0}                & $12_{\pm2}$          & $25_{\pm3}$          & $12_{\pm2}$  & $10_{\pm1}$  & $11_{\pm2}$          & $\mathbf{33_{\pm2}}$ \\
      \cmidrule{2-10}
      &\multirow{4}{*}{Puzzle}
                    & \multirow{4}{*}{Play}
                    & \texttt{visual-puzzle-3x3-play-v0}         & $0_{\pm0}$                                   & $\mathbf{21_{\pm1}}$ & $1_{\pm2}$           & $1_{\pm1}$   & $0_{\pm0}$   & $20_{\pm0}$                                 \\
                    &&                                            & \texttt{visual-puzzle-4x4-play-v0}           & $10_{\pm1}$          & $\mathbf{60_{\pm5}}$ & $16_{\pm4}$  & $0_{\pm0}$   & $10_{\pm6}$          & $26_{\pm2}$          \\
                    &&                                            & \texttt{visual-puzzle-4x5-play-v0}           & $5_{\pm2}$           & $17_{\pm1}$          & $7_{\pm2}$   & $0_{\pm0}$   & $6_{\pm1}$           & $\mathbf{20_{\pm0}}$ \\
                    &&                                            & \texttt{visual-puzzle-4x6-play-v0}           & $2_{\pm1}$           & $\mathbf{15_{\pm1}}$ & $2_{\pm1}$   & $0_{\pm0}$   & $3_{\pm1}$           & $7_{\pm0}$           \\
      \bottomrule
    \end{tabular}
  }
\end{table}

\begin{table}[htbp]
  \centering
  \scriptsize
  \setlength{\tabcolsep}{4pt}
  \caption{\textbf{Success rate ($\%$, $\uparrow$) on the effectiveness of noise steering
      and optimizer choices on OGBench datasets.} Results are mean $\pm$ standard deviation across
    8 (state) or 4 (pixel) seeds with the best result in bold; the first-order variant is \epssgd{} or \epsadam{} as in
    Table~\ref{tab:plan-dataset}, and $K=0$ is the reactive policy. In general, first-order methods
    are more effective on locomotion tasks, while zeroth-order methods are better on manipulation
    tasks. Noise steering allows the policy to improve itself at test time
    using imagined rollouts.}
  \label{tab:steering}

  \resizebox{\linewidth}{!}{
    \begin{tabular}{llccc}
      \toprule
      \textbf{Task Type}             & \textbf{Dataset}                             & \makecell{\method{}\\ (\epses{})} & \makecell{\method{} \\(\epssgd{} or \epsadam{})} & \makecell{\method{} \\ w/o steering ($K=0$)} \\
      \midrule
      \multirow{12}{*}{Locomotion}   & \texttt{pointmaze-medium-navigate-v0}        & $\mathbf{98\pm 1}$            & $\mathbf{98\pm 2}$                  & $97\pm 1$                      \\
                                     & \texttt{pointmaze-large-navigate-v0}         & $98\pm 1$            & $\mathbf{99\pm 1}$                  & $97\pm 1$                      \\
                                     & \texttt{pointmaze-giant-navigate-v0}         & $77\pm 3$            & $\mathbf{90\pm 3}$                  & $54\pm 3$                      \\
                                     & \texttt{pointmaze-teleport-navigate-v0}      & $60\pm 3$            & $\mathbf{64\pm 2}$                  & $57\pm 2$                      \\
                                     & \texttt{antmaze-medium-navigate-v0}          & $90\pm2$             & $\mathbf{96\pm 1}$                  & $83\pm 4$                      \\
                                     & \texttt{antmaze-large-navigate-v0}           & $75\pm 3$            & $\mathbf{85\pm 2}$                  & $70\pm 3$                      \\
                                     & \texttt{antmaze-giant-navigate-v0}           & $26\pm 2$            & $\mathbf{33\pm 3}$                  & $12\pm 1$                      \\
                                     & \texttt{antmaze-teleport-navigate-v0}        & $44\pm 5$            & $\mathbf{58\pm 3}$                  & $34\pm 2$                      \\
                                     & \texttt{visual-antmaze-medium-navigate-v0}   & $90\pm 2$            & $\mathbf{96\pm 1}$                  & $84\pm 1$                      \\
                                     & \texttt{visual-antmaze-large-navigate-v0}    & $70\pm 5$            & $\mathbf{84\pm 1}$                  & $64\pm 4$                      \\
                                     & \texttt{visual-antmaze-giant-navigate-v0}    & $0\pm 0$             & $\mathbf{10\pm 1}$                  & $5\pm 2$                       \\
                                     & \texttt{visual-antmaze-teleport-navigate-v0} & $40\pm 4$            & $\mathbf{54\pm 2}$                  & $33\pm 3$                      \\
      \midrule
      \multirow{18}{*}{Manipulation} & \texttt{cube-single-play-v0}                 & $93\pm 1$            & $\mathbf{98\pm 1}$                  & $26\pm 2$                      \\
                                     & \texttt{cube-double-play-v0}                 & $\mathbf{60\pm 1}$   & $55\pm 2$                           & $2\pm 1$                       \\
                                     & \texttt{cube-triple-play-v0}                 & $\mathbf{10\pm 2}$   & $5\pm 1$                            & $0\pm 0$                       \\
                                     & \texttt{cube-quadruple-play-v0}              & $\mathbf{1\pm 0}$    & $0\pm 0$                            & $0\pm 0$                       \\
                                     & \texttt{scene-play-v0}                       & $\mathbf{78\pm 2}$   & $64\pm 2$                           & $25\pm 2$                      \\
                                     & \texttt{puzzle-3x3-play-v0}                  & $\mathbf{93\pm 1}$   & $64\pm 5$                           & $1\pm 1$                       \\
                                     & \texttt{puzzle-4x4-play-v0}                  & $\mathbf{79\pm 4}$   & $57\pm 4$                           & $1\pm 1$                       \\
                                     & \texttt{puzzle-4x5-play-v0}                  & $\mathbf{20\pm 2}$   & $0\pm 0$                            & $0\pm 0$                       \\
                                     & \texttt{puzzle-4x6-play-v0}                  & $\mathbf{12\pm 1}$   & $0\pm 0$                            & $0\pm 0$                       \\
                                     & \texttt{visual-cube-single-play-v0}          & $\mathbf{83\pm 3}$   & $67\pm 1$                           & $56\pm 3$                      \\
                                     & \texttt{visual-cube-double-play-v0}          & $\mathbf{30\pm 1}$   & $27\pm 3$                           & $16\pm 2$                      \\
                                     & \texttt{visual-cube-triple-play-v0}          & $\mathbf{20\pm 1}$   & $17\pm 2$                           & $14\pm 2$                      \\
                                     & \texttt{visual-cube-quadruple-play-v0}       & $\mathbf{6\pm 0}$    & $3\pm 1$                            & $3\pm 2$                       \\
                                     & \texttt{visual-scene-play-v0}                & $\mathbf{33\pm 2}$   & $11\pm 2$                           & $6\pm 1$                       \\
                                     & \texttt{visual-puzzle-3x3-play-v0}           & $\mathbf{20\pm 0}$   & $0\pm 0$                            & $0\pm 0$                       \\
                                     & \texttt{visual-puzzle-4x4-play-v0}           & $\mathbf{26\pm 2}$   & $0\pm 0$                            & $0\pm 0$                       \\
                                     & \texttt{visual-puzzle-4x5-play-v0}           & $\mathbf{20\pm 0}$   & $0\pm 0$                            & $0\pm 0$                       \\
                                     & \texttt{visual-puzzle-4x6-play-v0}           & $\mathbf{7\pm 0}$    & $0\pm 0$                            & $0\pm 0$                       \\
      \bottomrule
    \end{tabular}
  }
\end{table}

\clearpage
\section{Discussion, Limitations, and Future Directions}
\label{app:discussion}

We discuss several limitations of \method{} and outline promising avenues for future research.

\subsection{Discussion}

\textbf{Trajectory smoothness and step efficiency.}
While \method{} reliably reaches specified goals across diverse environments, the resulting
trajectories are not always smooth or time-optimal. Because the closed-form Monte Carlo return
supervises the value function using hindsight temporal distances from suboptimal play trajectories,
the learned value landscape reflects the behavior distribution rather than shortest-path optimality.
Furthermore, receding-horizon replanning over discrete action chunks can introduce minor
discontinuities between consecutive plans. Devising new representations or abstractions for actions or shortest-path planning objectives to encourage completing tasks in the minimal number of steps remains an important next step.

\textbf{Advantage over-optimization under extended search.}
As shown in Section~\ref{sec:q-planner}, increasing the number of steering iterations $K$ generally
improves planning performance, yet we also notice that on certain datasets the success rate peaks and subsequently declines even as the imagined generalized advantage increases monotonically. This divergence indicates that aggressive test-time search may eventually exploit approximation errors in the learned dynamics and value heads rather than discovering better behaviors. Combining noise-space steering with uncertainty quantification or pessimistic penalties is a natural extension to prevent over-optimization.

\textbf{Short planning horizons.}
\method{} currently plans over a relatively short lookahead horizon (typically a single action chunk
of length $H=16$), relying on the terminal value estimate to measure longer-term progress towards goals. While effective when the learned value landscape is well-shaped, a short planning
horizon can limit performance on complex, multi-stage tasks where local value gradients are flat or
deceptive. Extending the effective lookahead to a horizon of \emph{hundreds} of steps without incurring prohibitive rollout variance or computational overhead is a valuable direction.

\textbf{Per-dataset hyperparameter tuning.}
Both training and test-time steering hyperparameters, such as learning rates, distillation weights, and planning iterations, are currently tuned per dataset
(Tables~\ref{tab:hparams-dataset} and~\ref{tab:plan-dataset}). Although this protocol reveals the
full capability of value-predictive planning across distinct locomotion and manipulation tasks,
manual per-dataset tuning is laborious. 

\subsection{Future Directions}

\textbf{Unified training across heterogeneous datasets.}
An appealing step towards eliminating per-dataset hyperparameter tuning is to train a single generalist value-predictive agent across heterogeneous datasets, \eg, combining diverse OGBench datasets such as exploratory play and noisy trajectories from different environments. Because closed-form hindsight return regression does not rely on
fragile temporal-difference bootstrapping or task-specific reward scales, it is well suited for
multi-task pre-training under a single, unified configuration.

\textbf{Scaling value-predictive world models.}
Our results demonstrate that compact, value-predictive world models already outperform
state-predictive counterparts on standard benchmarks. Investigating the scaling laws of
value-predictive representations, with respect to model capacity, dataset size, and task
diversity, is a compelling future direction to explore their limits.

\textbf{Synthesizing long-horizon experience at test time.}
Unlike Dyna-style methods that synthesize imagined rollouts to update the value function or policy
during training, \method{} focuses on planning within imagination at test time. Extending this
paradigm to generate reliable, long-horizon synthetic experience during deployment---for instance,
by chaining multi-chunk imagined rollouts or performing lightweight test-time adaptation and
self-improvement in imagination after offline pre-training---could bridge the gap between
short-horizon action-chunk steering and long-horizon reasoning.

%%%%%%%%%%%%%%%%%%%%%%%%%%%%%%%%%%%%%%%%%%%%%%%%%%%%%%%%%%%%

\end{document}